\documentclass{article} 
\usepackage{iclr2021_conference,times}

\usepackage{amsmath,amsfonts,bm}

\def\eqref#1{equation~\ref{#1}}

\def\1{\bm{1}}

\DeclareMathAlphabet{\mathsfit}{\encodingdefault}{\sfdefault}{m}{sl}
\SetMathAlphabet{\mathsfit}{bold}{\encodingdefault}{\sfdefault}{bx}{n}

\usepackage{hyperref}
\usepackage{url}
\usepackage[colorinlistoftodos]{todonotes}
\usepackage{xcolor}
\usepackage{svg}
\usepackage{subcaption}
\usepackage{booktabs}
\usepackage{multirow}
\usepackage{wrapfig}
\usepackage{ulem}
\usepackage{amsmath}
\usepackage[vlined,ruled]{algorithm2e}
\usepackage{floatrow}
\usepackage{blindtext}
\newfloatcommand{capbtabbox}{table}[][\FBwidth]
\newcommand{\mr}[1]{{\color{red}[\textbf{MR}:#1]}}

\title{Efficient Continual Learning with Modular Networks and Task-Driven Priors}

\author{Tom Veniat \\
LIP6, Sorbonne Université, France\\
\texttt{tom.veniat@lip6.fr} \\
\And
Ludovic Denoyer \& Marc'Aurelio Ranzato \\
Facebook Artificial Intelligence Research\\
\texttt{\{denoyer,ranzato\}@fb.com} \\
}
\iclrfinalcopy

\begin{document}

\maketitle

\begin{abstract}
Existing literature in Continual Learning (CL) has focused on
overcoming catastrophic forgetting, the inability of the learner to
recall how to perform tasks observed in the past.
There are however other desirable properties of a CL
system, such as the ability to transfer knowledge from previous tasks
and to scale memory and compute sub-linearly with the number
of tasks. Since most current benchmarks focus only on forgetting using short streams
of tasks, we first propose a new suite of benchmarks to probe CL algorithms  
across these new axes. Finally, we introduce a new modular architecture, whose modules represent atomic skills that can be composed to perform a certain task.
Learning a task reduces to figuring out which past modules to re-use, and which new modules to instantiate to solve the current task. Our learning algorithm leverages a task-driven prior over the exponential search space of all possible ways to combine modules, 
enabling efficient learning on long streams of tasks. 
Our experiments show that this modular architecture and learning algorithm perform competitively on widely used CL benchmarks while yielding superior performance on
the more challenging benchmarks we introduce in this work.
\end{abstract}

\section{Introduction}

Continual Learning (CL) is a learning framework whereby an agent learns
through a sequence of tasks~\citep{Ring1994, thrun-iros94, lifelong}, observing each task once and only once.
Much of the focus of the CL literature has been on avoiding
\textit{catastrophic forgetting}~\citep{hippocampus, catastrophic, catastrophic_empirical}, the inability of the learner to recall how to
perform a task learned in the past. In our view, remembering how to
perform a previous task is particularly important because it promotes
knowledge accrual and transfer.
 CL has then the potential to address one of
the major limitations of modern machine learning: its reliance on
large amounts of labeled data. An agent may learn well a new task
even when provided with little labeled data if it can leverage the knowledge accrued while
learning previous tasks. 

Our first contribution is then to pinpoint general properties that a
good CL learner should have, besides avoiding forgetting. In \textsection\ref{sec:evalCL}
we explain how the learner should be able to \textit{transfer knowledge} from related tasks
seen in the past. At the same time, the learner should be able to \textit{scale
sub-linearly} with the number of tasks, both in terms of memory and
compute, when these are related. 

\begin{figure}
    \vspace{-3em}
	\centering

	\begin{subfigure}[t]{0.48\linewidth}
		\centering
	    \includegraphics[width=\textwidth]{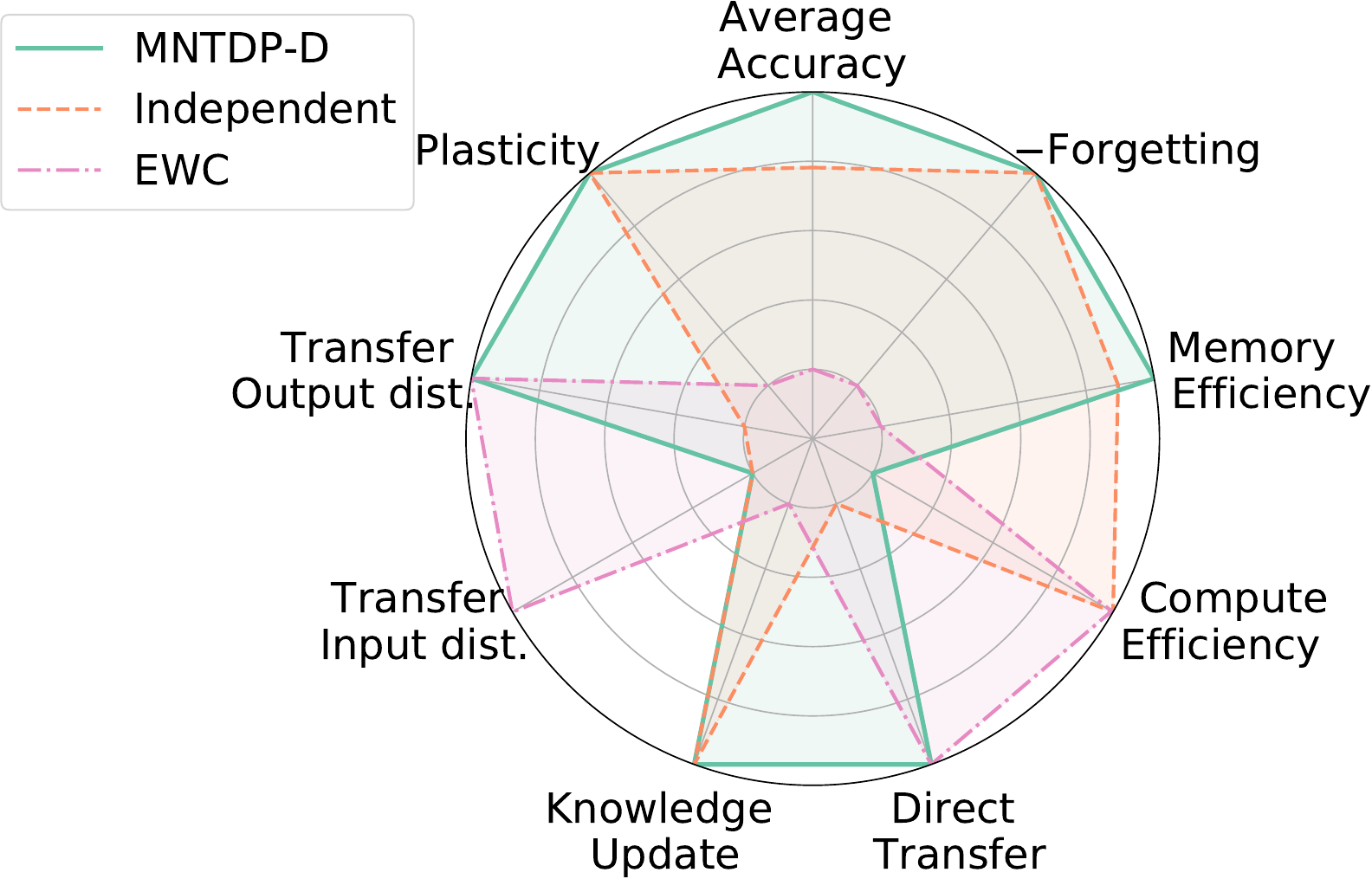}

	\end{subfigure}	
	\begin{subfigure}[t]{0.48\linewidth}
	  \centering	
	  \includegraphics[width=\textwidth]{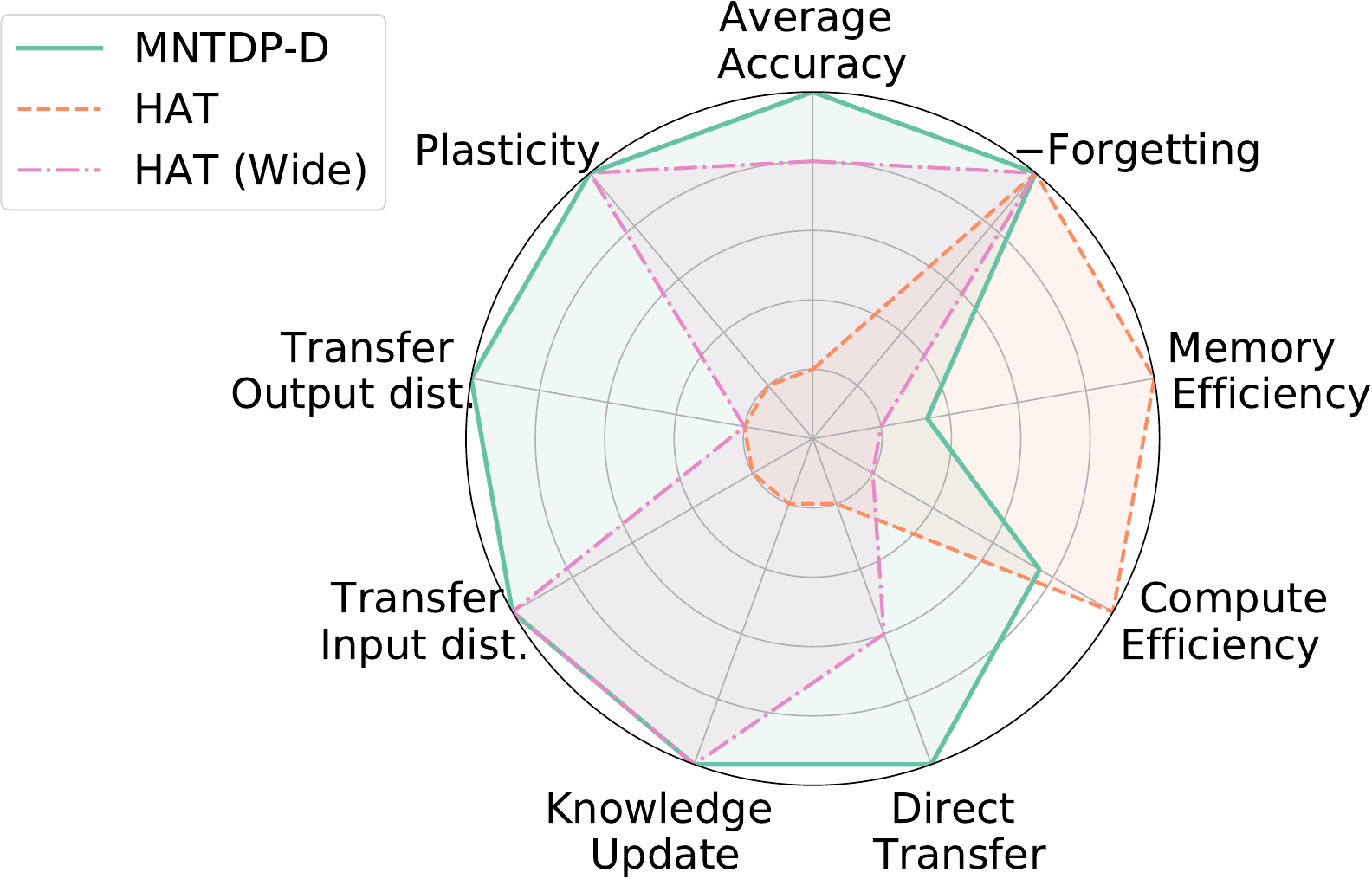}

	\end{subfigure}
    \caption{Comparison of various CL methods on the CTrL
      benchmark using Resnet (left) and Alexnet (right)
      backbones. MNTDP-D is our method. See
      Tab.~\ref{tab:res-all-metrics} of
      \textsection\ref{sec:old_bench} for details.}
    
        \label{fig:radar_chart_first_page}
\end{figure}

Our second contribution is to introduce a new benchmark suite, dubbed CTrL, to test
the above properties, since current benchmarks only focus on
forgetting. For the sake of simplicity and as a first step towards a
more holistic evaluation of CL models, in this work we restrict our
attention to supervised learning tasks and basic transfer learning
properties. Our experiments show that
while commonly used benchmarks do not discriminate well between
different approaches, our newly introduced benchmark let us dissect
performance across several new dimensions of transfer and scalability
(see Fig.~\ref{fig:radar_chart_first_page} for instance), 
helping machine learning developers better
understand the strengths and weaknesses of various approaches.

Our last contribution is a new model that is designed according to the
above mentioned properties of CL methods. It is based on a modular neural network
architecture~\citep{eigen14, denoyer15,
  DBLP:journals/corr/FernandoBBZHRPW17,DBLP:conf/icml/LiZWSX19} with a
novel task-driven prior (\textsection\ref{sec:mntdp}).
Every task is solved by the composition of a handful of
neural modules which can be either borrowed from past tasks or freshly
trained on the new task. In principle, modularization takes care of all the
fundamental properties we care about, as i) by design there is no forgetting as
modules from past tasks are not updated when new tasks arrive, ii)
transfer is enabled via sharing modules across tasks, and iii)
the overall model scales sublinearly with the number of tasks as long
as similar tasks share modules. The key issue is how to
efficiently select modules, as the search space grows exponentially in
their number. In this work, we overcome this problem by leveraging a \textit{data
driven prior} over the space of possible architectures, which
allows only local perturbations around the architecture
of the previous task whose features best solve the current task
(\textsection\ref{sec:ddp}).

Our experiments in~\textsection\ref{sec:experiments}, which employ a
stricter and more realistic evaluation protocol whereby streams are observed only once but
data from each task can be played multiple times, show that this model
performs at least as well as state-of-the-art methods on
standard benchmarks, and much better on our new and more challenging                             
benchmark, exhibiting better transfer and ability to scale to streams                  
with a hundred tasks.                                                                                          


\section{Related Work}
\label{sec:rw}
CL methods can be categorized into three main families of approaches.
{\bf Regularization} based methods use a single shared predictor
across all tasks with the only exception that there can be a task-specific 
classification head depending on the setting. They rely on various regularization methods to prevent forgetting.
 \citet{DBLP:journals/corr/KirkpatrickPRVD16,
   DBLP:conf/icml/Schwarz0LGTPH18} use an approximation of the Fisher
 Information matrix while \citet{DBLP:conf/icml/ZenkePG17} using the
 distance of each weight to its initialization as a measure of
 importance. These approaches work well in streams containing a
 limited number of tasks but will inevitably either forget or stop
 learning as streams grow in size and
 diversity~\citep{DBLP:journals/corr/abs-1904-07734}, due to their
 structural rigidity and fixed capacity.  

Similarly, {\bf rehearsal} based methods also share a single
 predictor across all tasks but attack forgetting by using rehearsal
 on samples from past tasks. For instance,
 \citet{DBLP:conf/nips/Lopez-PazR17,
   DBLP:journals/corr/abs-1902-10486, DBLP:conf/nips/RolnickASLW19}
 store past samples in a replay buffer, while
 \citet{DBLP:conf/nips/ShinLKK17} learn to generate new samples from
 the data distribution of previous tasks and
 \citet{DBLP:journals/corr/abs-1905-09447} computes per-class
 prototypes. These methods share the same drawback of regularization
 methods: Their capacity is fixed and pre-determined which makes them
 ineffective at handling long streams.

 Finally, approaches based on {\bf evolving architectures} directly
 tackle the issue of the limited capacity by enabling the architecture
 to grow over time.
 \citet{DBLP:journals/corr/RusuRDSKKPH16} introduce a new
 network on each task, with connection to all previous
 layers, resulting in a network that grows super-linearly with the
 number of tasks. This issue was later addressed
 by~\citet{DBLP:conf/icml/Schwarz0LGTPH18} who  propose to distill the new
 network back to the original one after each task, henceforth yielding
 a fixed capacity predictor which is going to have severe limitations on long streams.
\citet{DBLP:conf/iclr/YoonYLH18, DBLP:conf/nips/Hung0WCCC19} propose
a heuristic algorithm to automatically add and prune
weights. \citet{DBLP:conf/icml/LiZWSX19} propose to softly select between
reusing, adapting, and introducing a new module at every
layer. Similarly, \citet{RCL} propose to add filters once a new task
arrives using REINFORCE~\citep{reinforce}, leading to larger and
larger networks even at inference time as time goes by. These two
works are the most similar to ours, with the major difference that we
restrict the search space over architectures, enabling much better scaling
to longer streams. While their search space (and RAM consumption) grows over time, ours is
constant. Our approach is \textit{modular}, and only a small (and
constant) number of modules is employed for any given
task both at training and test time. Non-modular approaches, like those relying on individual neuron
gating~\citep{claw, HAT, IBP}, lack such runtime efficiency which limits their
applicability to long streams. 
\citet{DBLP:journals/corr/abs-1906-01120, DBLP:journals/corr/FernandoBBZHRPW17} propose
to learn a modular architecture, where each task is identified by a path
in a graph of modules like we do. However, they lack the prior over
the search space. They both use random search 
which is rather inefficient as the number of modules grows.

There are of course other works that introduce new benchmarks for
CL. Most recently, \citet{DBLP:journals/corr/abs-2006-14769} have
proposed a stream with 2500 tasks all derived from MNIST
permutations. Unfortunately, this may provide little insight in terms
of how well models transfer knowledge across tasks. Other benchmarks
like CORe50~\citep{core50} and CUB-200~\citep{cub200} are more realistic
but do not enable precise assessment of how well models transfer and scale.


CL is also related to other learning paradigms, such as
meta-learning~\citep{DBLP:conf/icml/FinnAL17,
  DBLP:journals/corr/abs-1803-02999, DBLP:journals/corr/DuanSCBSA16},
but these only consider the problem of quickly adapting to a new task
while in CL we are also interested in preventing forgetting and
learning better over time.  For instance,
\citet{DBLP:conf/corl/AletLK18} proposed a modular approach for
robotic applications. However, only the performance on the last task
was measured. There is also a body of literature on modular networks
for multi-task and multi-domain
learning~\citep{DBLP:conf/aaai/RuderBAS19, DBLP:conf/nips/RebuffiBV17,
  DBLP:journals/corr/abs-2007-12415}. The major differences are the
static nature of the learning problem they consider and the lack of
emphasis on scaling to a large number of tasks.

\section{Evaluating Continual Learning Models} \label{sec:evalCL}
Let us start with a general formalization of the CL
framework. We assume that tasks arrive in sequence, and that each
task is associated with an integer task descriptor $t=1,2,...$ which
corresponds to the order of the tasks. Task descriptors are provided
to the learner both during training and test time.
Each task is defined by a labeled dataset $\mathcal{D}^t$.
We denote with $\mathcal{S}$ a sequence of such tasks. A predictor for a given task $t$ is denoted by $f: \mathcal{X}^{t}
\times \mathbb{Z} \rightarrow \mathcal{Y}^{t}$. 
The predictor has internally some trainable
parameters whose values depend on the stream of tasks $\mathcal{S}$ seen in the past,
therefore the prediction is: $f(x, t | \mathcal{S})$. Notice that in
general, different streams lead to different predictors for the same
task: $f(x, t | \mathcal{S}) \neq f(x, t | \mathcal{S}')$.


\paragraph{Desirable Properties of CL models And Metrics: } \label{sec:metrics}
Since we are focusing on supervised learning tasks, it is natural to
evaluate models in terms of accuracy. We denote the prediction accuracy of
the predictor $f$ as $\Delta( f(x,
t | \mathcal{S}), y)$, where $x$ is the input, $t$ is the task
descriptor of $x$, $ \mathcal{S}$ is the stream of tasks seen by the
learner and $y$ is the ground truth label.

In this work, we consider four major properties of a CL algorithm.
First, the algorithm has to yield predictors that are accurate by the
end of the learning experience. This is measured by their {\bf average
accuracy} at the end of the learning experience:
\begin{equation}
  \mathcal{A}(\mathcal{S}) = \frac{1}{T}\sum\limits_{t=1}^{T}
  \mathbb{E}_{(x,y) \sim \mathcal{D}^t} [
 \Delta(f(x, t |\mathcal{S}= 1, \dots, T), y)]. \label{eq:avgacc}
\end{equation}
Second, the CL algorithm should
yield predictors that do not forget, i.e. that are able to
perform a task seen in the past without significant loss of
accuracy. {\bf Forgetting} is defined as:
\begin{equation}
\mathcal{F}(\mathcal{S}) = \frac{1}{T-1} \sum_{t=1}^{T-1} \mathbb{E}_{(x,y) \sim \mathcal{D}^t} [
\Delta(f(x, t | \mathcal{S} = 1, \dots, T), y ) - \Delta(f(x, t|
\mathcal{S}' = 1, \dots, t), y) ]
  \label{eq:forgetting}
\end{equation}
This measure of forgetting has been called backward
transfer~\citep{DBLP:conf/nips/Lopez-PazR17}, and it measures the average loss of accuracy on
a task at the end of training compared to when the task was just
learned. Negative values indicate the model has been
forgetting. Positive values indicate that the model has been improving
on past tasks by learning subsequent tasks.

Third, the CL algorithm should yield predictors that are capable of
transferring knowledge from past tasks when solving a new task. {\bf Transfer}
 can be measured by:
 \begin{equation}
  \mathcal{T}(\mathcal{S})= \mathbb{E}_{(x,y) \sim \mathcal{D}^T}
  [\Delta(f(x, T |  \mathcal{S} = 1, \dots, T), y) -
  \Delta(f(x, T |  \mathcal{S}' = T), y)] \label{eq:transfer}
 \end{equation}
which measures the difference of performance between a model
that has learned through a whole sequence of tasks and a model that 
has learned the last task in isolation.
 We would expect this
quantity to be positive if there exist previous tasks that are related
to the last task. Negative values imply the model has suffered some
form of interference or even lack of plasticity when the
predictor has too little capacity left to learn the new task.

Finally, the CL algorithm has to yield predictors that {\bf scale
sublinearly} with the number of tasks both in terms of memory and
compute. In order to quantify this, we simply report the total memory
usage and compute by the end of the learning experience during
training.
We therefore include in the memory consumption everything a learner has to keep
around to be able to continually learn (e.g., regularization
parameters of EWC or the replay buffer for experience replay).

\paragraph{Streams} \label{sec:streams}
The metrics introduced above can be applied to any
stream of tasks. While current benchmarks are constructed to assess
forgetting, they fall short at enabling a comprehensive evaluation of
\textit{transfer} and \textit{scalability} because they do not
control for task relatedness and they are composed of too few tasks.
Therefore, we propose a new suite of streams. If
 $t$ is a task in the stream, we denote with $t^-$ and $t^+$  a task whose data
is sampled from the same distribution as $t$, but with a much smaller
or larger labeled dataset, respectively. Finally, $t'$ and $t''$ are tasks that
are similar to task $t$, while we assume no relation between $t_i$ and
$t_j$, for all $i \neq j$.

We consider five axes of transfer 
and define a stream for each of them. While other dimensions certainly
exist, here we are focusing on basic properties that any desirable
model should possess. 

\textbf{Direct Transfer}: we define the stream $\mathcal{S}^{\mbox{\small{-}}} = (t_1^+, t_2, t_3, t_4, t_5, t_1^-)$ where the last task is the same as the first one
but with much less data to learn from. This is useful to assess
whether the learner can directly transfer
knowledge from the relevant task.\\
\textbf{Knowledge Update}: we define the stream $\mathcal{S}^{\mbox{\small{+}}} = (t_1^-, t_2, t_3, t_4, t_5, t_1^+)$ where the last
task has much more data than the first task with intermediate tasks
that are unrelated. In this case, there should not be much need to transfer
anything from previous tasks, and the system can just use the last
task to update its knowledge of the first task.\\ 
\textbf{Transfer to similar Input/Output Distributions}: we define two streams where the
last task is similar to the first task but the input distribution changes 
$\mathcal{S}^{\mbox{\tiny{in}}} = (t_1, t_2, t_3, t_4, t_5, t_1')$, or the output
distribution changes $\mathcal{S}^{\mbox{\tiny{out}}} = (t_1, t_2, t_3, t_4,
t_5, t_1'')$.\\
\textbf{Plasticity:} this is a stream where all tasks are unrelated,  $\mathcal{S}^{\mbox{\tiny{pl}}}
= (t_1, t_2, t_3, t_4, t_5)$, which is useful to measure the 
''ability to still learn'' and potential interference (erroneous transfer from unrelated tasks) when learning the last task. 

All these tasks are evaluated using $\mathcal{T}$ in
eq.~\ref{eq:transfer}. Other dimensions of transfer (e.g., transfer
with compositional task descriptors or under noisy conditions) are
avenues of future work. Finally, we evaluate scalability using 
$\mathcal{S}^{\mbox{\tiny{long}}}$, a stream with 100 tasks of various degrees of
relatedness and with varying amounts of training data. See
\textsection\ref{sec:CTrL} for more details.

\section{Modular Networks with Task Driven Prior (MNTDP)} \label{sec:mntdp}
\begin{figure}[t]
  \centering
  \includegraphics[width=.8\linewidth]{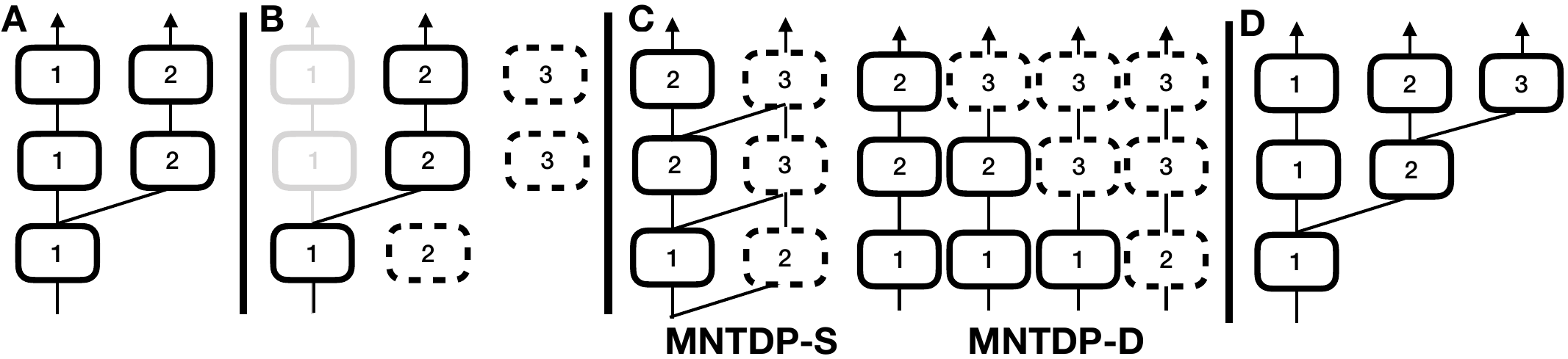}
  \vspace{-.2cm}
   \caption{\small Toy illustration of the approach when each
     predictor is composed of only three modules and
     only two tasks have already been observed. {\bf A)}: The
     predictor of the first task uses modules (1,1,1) (listing modules
     by increasing depth in the network) while the
     predictor of the second task uses modules (1,2,2); the first
     layer module is shared between the two predictors.
     {\bf B)}: When a new task arrives, first we add one
     new randomly initialized module at each layer (the dashed
     modules). Second, we search for the most similar past task and
     retain only the corresponding architecture. In this case, the
     second task is most similar and therefore we remove (gray out)
     the modules used only by the predictor of the first task. {\bf
       C)}: We train on the current task by learning both the best way
     to combine modules and their parameters. However, we restrict the
     search space. In this
     case, we only consider four possible compositions, all derived by
     perturbing the predictor of the second task. In the stochastic
     version (MNTDP-S), for every input a path (sequence of modules)
     is selected stochastically. Notice that the same module may
     contribute to multiple paths (e.g., the top-most layer with id 3).
     In the deterministic version instead (MNTDP-D), we train in
     parallel all paths and then select the best. Note that only the
     parameters of the newly added (dashed) modules are subject to learning.
     {\bf D)}: Assuming that the best architecture found at the
     previous step is (1,2,3), module 3 at the top layer is added to
     the current library of modules.}
   \vspace{-.2cm}
  \label{fig:toy_illustration}
\end{figure}

In this section we describe an approach, dubbed MNTDP,  designed according
to the properties discussed in \textsection\ref{sec:metrics}.
 The class of predictor functions $f(x, t | \mathcal{S})$ we consider in this work is \textit{modular},
in the sense that predictors are composed of modules that are potentially
shared across different (but related) tasks. A module can be any parametric
function, for instance, a ResNet block \citep{resnet}. The only restriction we use in
this work is that all modules in a layer should differ only in terms of their actual
parameter values, while modules across layers can implement different
classes of functions. For instance, in the toy
illustration of Fig.~\ref{fig:toy_illustration}-A (see caption for
details), there are two
predictors, each composed of three modules (all ResNet blocks), the first one being shared.

\subsection{Training} \label{sec:training}
Once the new
task $t$ arrives, the algorithm follows three
steps. First, it temporarily adds new randomly initialized modules at
every layer (these are denoted by dashed boxes in
Fig.~\ref{fig:toy_illustration}-B) and it
then defines a search space over all possible ways to combine 
modules. Second, it minimizes a loss function, which in our case is
the cross-entropy loss, over
both ways to combine modules and module parameters, see
Fig.~\ref{fig:toy_illustration}-C. Note that only the parameters of the newly
added modules are subject to training, de facto preventing forgetting
of previous tasks by construction but also preventing positive
backward transfer. 
Finally, it takes
the resulting predictor for the current task and it adds the
parameters of the newly added modules (if any) back to the existing
library of module parameters, see Fig.~\ref{fig:toy_illustration}-D.

Since predictors are uniquely identified by which modules compose
them, they can also be described by the \textit{path} in the grid of
module parameters. We denote the $j$-th path in the graph by $\pi_j$. The parameters of the
modules in path $\pi_j$ are denoted by $\theta(\pi_j)$. Note that in
general $\theta(\pi_j) \cap \theta(\pi_i) \neq \emptyset$, for $i \neq
j$ since some modules may be shared across two different paths.

Let $\Pi$ be the set of all possible paths in the graph. This has a size
equal to the product of the number of modules at every layer, after
adding the new randomly initialized modules. If $\Gamma$ is a
distribution over $\Pi$ which is subject to learning (and initialized
uniformly), then the loss function is:
\begin{equation}
\Gamma^*, \theta^* =  \arg \min_{\theta, \Gamma} \mathbb{E}_{j \sim
  \Gamma, (x,y) \sim \mathcal{D}^t} \mathcal{L}( f(x, t| \mathcal{S}, \theta(\pi_j)), y) \label{eq:loss_general}
\end{equation}
where $f(x, t| \mathcal{S}, \theta(\pi_j)$ is the predictor using
parameters $\theta(\pi_j)$, 
$\mathcal{L}$ is the loss, and the minimization over the
parameters is limited to only the newly introduced modules.
The resulting distribution $\Gamma^*$
is a delta distribution, assuming no ties between paths. Once the best
path has been found and its parameters have been learned, the
corresponding parameters of the new modules in the optimal path are
added to the existing set of modules while the other ones are disregarded
(Fig.~\ref{fig:toy_illustration}-D). In this work, we consider two instances of the learning problem in
eq.~\ref{eq:loss_general}, which differ in a) how they optimize over
paths and b) how they share parameters across modules.

\paragraph{Stochastic version:} This algorithm alternates between one
step of gradient descent over the paths via
REINFORCE~\citep{reinforce} as in \citep{our_cvpr}, and one step of gradient
descent over the parameters for a given path.
The distribution $\Gamma$ is modeled by a product of multinomial
distributions, one for each layer of the model. These select one
module at each layer, ultimately determining a particular path.
Newly added modules may be
shared across several paths which
yields a model that can support several predictors while retaining a
very parsimonious memory footprint thanks to parameter sharing. This
version of the model, dubbed MNTDP-S, is outlined in the left part of
Fig.~\ref{fig:toy_illustration}-C and in
Alg.~\ref{algo:MNTDP-S} in the Appendix. In order to encourage the
model to explore multiple paths, we use an entropy regularization
on $\Gamma$ during training.

\paragraph{Deterministic version:} This algorithm minimizes the
objective over paths in eq.~\ref{eq:loss_general} via exhaustive
search, see Alg.~\ref{algo:MNTDP-D} in Appendix and the right part of
Fig.~\ref{fig:toy_illustration}-C. Here, paths do not share any
newly added module and we train one independent 
network per path, and then select the path yielding the lowest loss
on the validation set. While this requires much more memory, it may also
lead to better overall performance because each new module is cloned and
trained just for a single path. Moreover, training predictors on each
path can be trivially and cheaply parallelized on modern GPU devices.

\subsection{Data-Driven Prior}
\label{sec:ddp}
Unfortunately, the algorithms as described above do not scale to a
large number of tasks (and henceforth modules) because the search
space grows exponentially. This is also the case for other evolving
architecture approaches proposed in the
literature~\citep{DBLP:conf/icml/LiZWSX19,
  DBLP:journals/corr/abs-1906-01120, DBLP:conf/iclr/YoonYLH18, RCL} as
discussed in \textsection\ref{sec:rw}.

If there were $N$ modules per layer and $L$
layers, the search space would have size $N^L$. In order to restrict
the search space, we only allow paths that branch to the right: A newly
added module at layer $l$ can only connect to another newly added
module at layer $l+1$, but it cannot connect to an already trained
module at layer $l+1$. The underlying assumption is that for most
tasks we expect changes in the output distribution as opposed to the input
distribution, and therefore if tasks are related, the base
trunk is a good candidate for being shared. We will see in \textsection\ref{sec:old_bench} what happens when this assumption is not satisfied, e.g., when applying this to  $\mathcal{S}^{\mbox{\tiny{in}}}$.

To further restrict the search space we employ a \textit{data-driven
  prior}. The intuition is to limit the search space
to perturbations of the path corresponding to the past task (or to the
top-$k$ paths) that is the
most similar to the current task. There are several methods to assess
which task is the closest to the current task without accessing data
from past tasks and also different ways to perturb a path. We
propose a very simple approach, but others could have been used.
We take the predictors from all the past tasks and select the path that yields the best nearest neighbor classification
accuracy when feeding data from
the current task using the features just before the classification
head. This process is shown in Fig.~\ref{fig:toy_illustration}-B.
The search space is
reduced from $T^L$ to only $L$, and $\Gamma$ of
eq.~\ref{eq:loss_general} is allowed to have non-zero support only in this
restricted search space, yielding a much lower computational and
memory footprint which is \textit{constant} with respect to the number
of tasks. The designer of the model has now direct control (by varying
$k$, for instance) over the trade-off between accuracy and computational/memory budget.

By construction, the model does not forget, 
 because we do not update modules of previous tasks.
The model can transfer well because it can re-use modules from
related tasks encountered in the past while not being constrained in
terms of its capacity. And finally, the model scales
sub-linearly in the number of tasks because modules can be shared
across similar tasks. We will validate empirically in \textsection\ref{sec:old_bench}
whether the choice of the restricted search space works well in practice.


\section{Experiments} \label{sec:experiments}
In this section we first introduce our benchmark in
\textsection\ref{sec:CTrL} and the modeling details
\textsection\ref{sec:exp_modeling}, and then report results both on
standard benchmarks as well as ours in \textsection\ref{sec:old_bench}.\footnote{CTrL source code available at \url{https://github.com/facebookresearch/CTrLBenchmark}. 

Source code of the experiments available at \url{https://github.com/TomVeniat/MNTDP}.}

\subsection{The CTrL Benchmark} \label{sec:CTrL}
The CTrL (Continual Transfer Learning) benchmark is a collection of
streams of tasks  built over
seven popular computer vision datasets, namely: CIFAR10 and CIFAR100~\citep{cifar10},
 DTD~\citep{cimpoi14describing}, SVHN~\citep{svhn}, MNIST~\citep{mnist}, Rainbow-MNIST~\citep{rmnist}
and Fashion MNIST~\citep{xiao2017/online}; see Table
\ref{tab:tasks-datasets} in Appendix for basic statistics. These datasets are
desirable because they are diverse (hence tasks derived from some of these
datasets can be considered unrelated), they have a fairly large number
of training examples to simulate tasks that do not need to transfer,
and they have low spatial resolution enabling fast evaluation.
CTrL is designed according to the methodology described in
\textsection\ref{sec:streams}, to enable evaluation of various
transfer learning properties and the ability of models to scale to a large
number of tasks. Each task consists of a training, validation,
and test datasets corresponding to 
a 5-way and 10-way classification problem for the transfer streams and
the long stream, respectively. The last task of $\mathcal{S}^{\mbox{\tiny{out}}}$
 consists of a shuffling of the output
labels of the first task. The last task of $\mathcal{S}^{\mbox{\tiny{in}}}$ is the same as its
first task except that MNIST images have a different background
color. $\mathcal{S}^{\mbox{\tiny{long}}}$ has 100 tasks, and it is constructed by first sampling a
dataset, then 5 classes at random, and finally the amount of training
data from a distribution that favors small tasks by the end of the
learning experience. See Tab.~\ref{tab:long-stream-tasks-p1} and \ref{tab:long-stream-tasks-p2} in Appendix for details. Therefore,
$\mathcal{S}^{\mbox{\tiny{long}}}$ tests not only the ability of a model to scale to a
relatively large number of tasks but also to transfer more efficiently
with age. 

All images have been rescaled to 32x32 pixels in
RGB color format, and per-channel normalized using statistics computed on
the training set of each task.
During training, we perform data augmentation by using random crops (4 pixels padding and 32x32 crops) and random horizontal
reflection.
Please, refer to Appendix~\ref{app:datasets} for further
details.

\subsection{Methodology and Modeling Details} \label{sec:exp_modeling}

Models learn over each task in sequence; data from each task can be
replayed several times but \textit{each stream is observed only once}. 
Since each task has a validation dataset, hyper-parameters
(e.g., learning rate and number of weight updates) are
task-specific and they are cross-validated on the validation
set of each task. Once the learning experience ends, we test the resulting
predictor on the test sets of all the tasks. Notice that this is a
stricter paradigm than what is usually employed in the literature~\citep{DBLP:journals/corr/KirkpatrickPRVD16}, where
hyper-parameters are set at the stream level (by replaying the stream
several times). Our model selection criterion is more realistic
because it does not assume that the learner has access to future tasks
when cross-validating on the current task, and this is more
consistent with the CL's assumptions of operating on a stream of data.

All models use the same backbone. Unless otherwise specified, this is a small variant of the ResNet-18 architecture which is
divided into $7$ modules; please, refer to Table \ref{tab:resnet-arch-details} for
details of how many layers and parameters each block contains. Each
predictor is trained by minimizing the cross-entropy loss with a small
L2 weight decay on the parameters. In our experiments, MNTDP adds $7$
new randomly initialized modules, one for every block. The search
space does not allow connecting old blocks from new blocks, and it
considers two scenarios: using old blocks from the past task
that is deemed most similar ($k=1$, the default setting) or considering the whole set of
old blocks ($k=\textrm{all}$) resulting in a much larger search
space. 

We compare to several baselines: \textbf{Independent Models} which
instantiates a randomly initialized predictor for every task (as many
paths as tasks without any module overlap), 2) \textbf{Finetuning} which
trains a single model to solve all the tasks without any
regularization and initializes from the model of the previous task (a
single path shared across all tasks), 3) \textbf{New-Head} which also shares the
trunk of the network across all tasks but not the classification head
which is task-specific, 4)  \textbf{New-Leg} which shares all layers
across tasks except for the very first input layer which is task-specific, 5) \textbf{EWC}~\citep{DBLP:journals/corr/KirkpatrickPRVD16}
which is like ``finetuning'' but with a regularizer to alleviate
forgetting, 6)  \textbf{Experience Replay}~\citep{DBLP:journals/corr/abs-1902-10486} which is like finetuning
except that the model has access to some samples from the past tasks
to rehearse and alleviate forgetting (we use $15$ samples per class to obtain a memory consumption similar to other baselines), 7) \textbf{Progressive Neural
  Networks} (PNNs)~\citep{DBLP:journals/corr/RusuRDSKKPH16} which adds both a
new module at every layer as well as lateral connections once a new
task arrives.
8) \textbf{HAT}~\citep{HAT}: which learns an attention mask over the
parameters of the backbone network for each task. Since HAT's
open-source implementation uses AlexNet~\citep{Alexnet}
as a backbone, we also implemented a version of MNTDP using
AlexNet for a fair comparison. Moreover, we considered two versions of
HAT, the default as provided by the authors and a version, dubbed HAT-wide, that is as
wide as our final MNTDP model (or as wide as we can fit into GPU memory).
9) \textbf{DEN}~\citep{DBLP:conf/iclr/YoonYLH18} and 10)
\textbf{RCL}~\citep{RCL} which both evolve architectures. For these
two models since there is no publicly available implementation with
CNNs, we only report their performance on Permuted-MNIIST using fully
connected nets (see Fig.~\ref{fig:res-permmnist-barplot} and
Tab.~\ref{tab:res-mnist-perm} in Appendix~\ref{app:classic_streams}).

We report performance across different axes as discussed in \textsection\ref{sec:evalCL}: Average accuracy
as in eq.~\ref{eq:avgacc}, forgetting as in eq.~\ref{eq:forgetting},
transfer as in eq.~\ref{eq:transfer} and applicable only to the
transfer streams, ``Mem.'' [MB] which refers to
the average memory consumed by the end of training, and
``Flops'' [T] which corresponds to the average amount of computation used by the end of training.



\subsection{Results} \label{sec:old_bench}

\begin{figure}	
	\centering
	\begin{subfigure}[t]{0.49\linewidth}
		\centering
		\includegraphics[width=\linewidth]{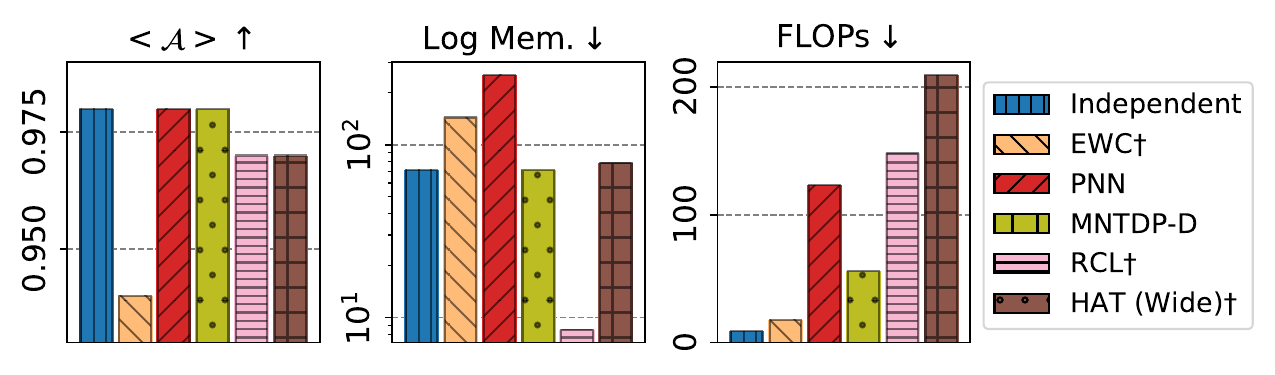}
		\vspace*{-7mm}
		\caption{Permuted-MNIST} \label{fig:res-permmnist-barplot}		
	\end{subfigure}	
	\captionsetup[subfigure]{oneside,margin={0cm,2cm}}
	\begin{subfigure}[t]{0.49\linewidth}
	  \centering
        \includegraphics[width=\linewidth]{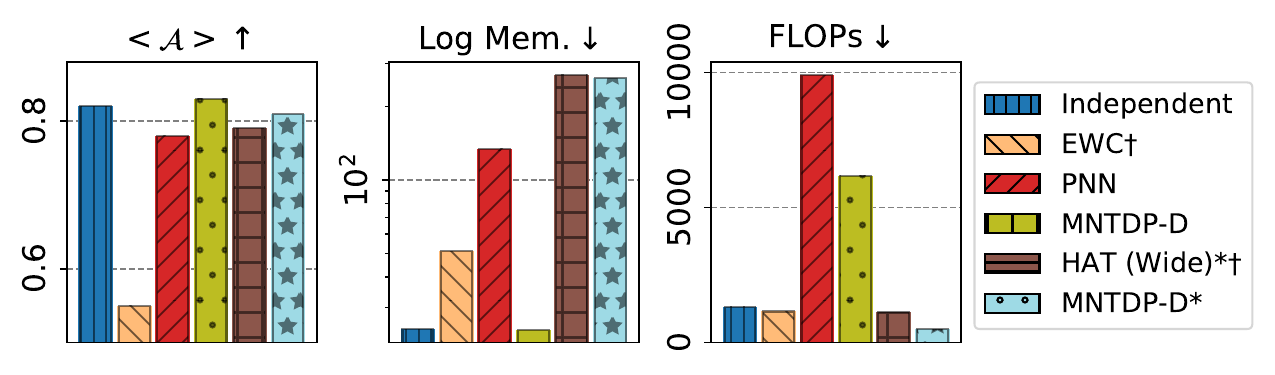}
        	\vspace*{-7mm}
		\caption{Split Cifar-100}\label{fig:res-splitcifar-barplot}		
	\end{subfigure}
	\vspace*{-3mm}
	\caption{\small Results on standard continual learning streams. * denotes an Alexnet Backbone. $\dagger$
          correspond to models cross-validated at the stream-level, a setting that favors them over the other methods which are cross-validated at the task-level. Detailed results are presented in          Appendix~\ref{app:classic_streams}.}
        \label{fig:old-bench-barplot}
\end{figure}

\paragraph{Existing benchmarks: }\label{sec:suppmat_standard_benchmarks}
In Fig.~\ref{fig:old-bench-barplot} we compare MNTDP against several baselines on two standard streams with 10
tasks, Permuted MNIST, and Split CIFAR100. 
We
observe that all models do fairly well, with EWC falling a bit behind
the others in terms of average accuracy.
PNNs has good average accuracy but
requires more memory and compute.
Compared to MNTDP, both RCL and HAT have lower average accuracy and require
more compute.
MNTDP-D yields the best average accuracy, but
requires more computation than ``independent models''; notice however that its wall-clock training
time is actually the same as ``independent models'' since all candidate paths (seven in our case) can be trained in parallel on
modern GPU devices. In fact, it turns out that on these standard streams MNTDP trivially reduces to
``independent models'' without any module sharing, since each task is 
fairly distinct and has a relatively large amount of data. It
is therefore not possible to assess how well models can transfer knowledge
across tasks, nor it is possible to assess how well models
scale.  Fortunately, we can leverage the CTrL benchmark to better
study these properties, as described next.

\paragraph{CTrL: } \label{sec:results_ctrl}
We first evaluate models in terms of their ability to transfer by
evaluating them on the streams $\mathcal{S}^{\mbox{\small{-}}},
\mathcal{S}^{\mbox{\small{+}}}, \mathcal{S}^{\mbox{\tiny{in}}}, \mathcal{S}^{\mbox{\tiny{out}}}$ and 
$\mathcal{S}^{\mbox{\tiny{pl}}}$ introduced in 
\textsection\ref{sec:streams}. Tab.~\ref{tab:res-all-metrics}
shows that ``independent models'' is again a strong baseline, because
even on the first four streams, all tasks except the last one are
unrelated and therefore instantiating an independent model is
optimal. However, MNTDP yields the best average accuracy overall. MNTDP-D
achieves the best transfer on streams $\mathcal{S}^{\mbox{\small{-}}},
\mathcal{S}^{\mbox{\small{+}}}, \mathcal{S}^{\mbox{\tiny{out}}}$ and
$\mathcal{S}^{\mbox{\tiny{pl}}}$, and indeed it discovers the correct path in
each of these cases (e.g., it discovers to reuse the path of the first
task when learning on $\mathcal{S}^{\mbox{\small{-}}}$ and to just swap the top modules
when learning the last task on $\mathcal{S}^{\mbox{\small{+}}}$). Examples of discovered path are presented in appendix~\ref{app:discovered_paths}. MNTDP underperforms on
$\mathcal{S}^{\mbox{\tiny{in}}}$ because its prior does not match the data
distribution, since in this case, it is the input distribution that has
changed but swapping the first module is out of MNTDP search
space. This highlights the importance of the choice of prior for this
algorithm. In general, MNTDP offers a clear trade-off between
accuracy, i.e. how broad the prior is which determines how many paths can be evaluated, and memory/compute budget.
Computationally MNTDP-D is the most demanding, but in
practice its wall clock time is comparable to ``independent'' because GPU devices can
store in memory all the paths (in our case, seven) and efficiently train them all in
parallel. We observe also that  MNTDP-S
has a clear advantage in terms of compute at the expense of a lower
overall average accuracy, as sharing modules across paths during
training can lead to sub-optimal
convergence.

\begin{table}[t]
  \centering
  \small
\begin{tabular}{p{2cm}p{1cm}p{1cm}p{0.7cm}p{0.7cm}p{0.7cm}p{0.7cm}p{0.7cm}p{0.7cm}p{1cm}}

\toprule
{}  &  $<\mathcal{A}>$ &  $<\mathcal{F}>$ &   Mem. &  FLOPs &  $\mathcal{T}(\mathcal{S}^{\mbox{\small{-}}})$ &  $\mathcal{T}(\mathcal{S}^{\mbox{\small{+}}})$ &  $\mathcal{T}(\mathcal{S}^{\mbox{\tiny{in}}})$ &  $\mathcal{T}(\mathcal{S}^{\mbox{\tiny{out}}})$ &  $\mathcal{T}(\mathcal{S}^{\mbox{\tiny{pl}}})$ \\
\midrule
Independent        &             0.58 &              \bf{0.0} &   14.1 &    308 &                           0.0 &                           \bf{0.0} &                                     0.0 &                                      0.0 &                                     \bf{0.0} \\
Finetune           &             0.19 &             -0.3 &    \bf{2.4} &    \bf{284} &                           0.0 &                          -0.1 &                                    -0.0 &                                     -0.0 &                                    -0.1 \\
New-head    &             0.48 &              \bf{0.0} &    2.5 &    307 &                           \bf{0.4} &                          -0.3 &                                    -0.2 &                                      \bf{0.3} &                                    -0.4 \\
New-leg     &             0.41 &              \bf{0.0} &    2.5 &    366 &
0.3 &                          -0.3 &
{\bf 0.4} &                                     -0.1 &                                    -0.4 \\
Online EWC $\dagger$        &             0.43 &             -0.1 &    7.3 &    310 &                           0.3 &                          -0.3 &                                     0.3 &                                      \bf{0.3} &                                    -0.4 \\
ER              &             0.44 &             -0.1 &   13.1 &    604 &                           0.0 &                          -0.2 &                                     0.0 &                                      0.1 &                                    -0.2 \\
PNN                &             0.57 &              \bf{0.0} &   48.2 &   1459 &                           0.3 &                          -0.2 &                                     0.1 &                                      0.2 &                                    -0.1 \\
MNTDP-S            &             0.59 &              \bf{0.0} &   11.7 &    363 &                           \bf{0.4} &                          -0.1 &                                     0.0 &                                      \bf{0.3} &                                    -0.1 \\
MNTDP-D            &             \bf{0.64} &              \bf{0.0} &   11.6 &   1512 &                           \bf{0.4} &                           \bf{0.0} &                                     0.0 &                                      \bf{0.3} &                                    \bf{-0.0} \\
\hline
MNTDP-D*  &             0.62 &              0.0 &  140.7 &    115 &                           0.3 &                          -0.1 &                                     0.1 &                                      0.3 &                                    -0.1 \\
HAT*$\dagger$      &             0.58 &             -0.0 &   26.6 &     45 &                           0.1 &                          -0.2 &                                     0.0 &                                      0.1 &                                    -0.2 \\
HAT (Wide)*$\dagger$ &             0.61 &              0.0 &  163.9 &    274 &                           0.2 &                          -0.1 &                                     0.1 &                                      0.1 &                                    -0.1 \\
\bottomrule
\end{tabular}

    \caption{\label{tab:res-all-metrics} \small Aggregated results on the
      transfer streams over multiple relevant baselines (complete table with more baselines provided in Appendix~\ref{app:detailed_results}). * correspond to models using an Alexnet backbone.}
\end{table}

Overall, MNTDP has a much higher average accuracy than methods
with a fixed capacity. It also beats PNNs, which seems to struggle
with interference when learning on $\mathcal{S}^{\mbox{\tiny{pl}}}$, as all
new modules connect to all old modules which are irrelevant for the
last task. Moreover, PNNs uses much more memory.
 ``New-leg'' and ``new-head'' models perform very well only
 when the tasks in the stream match their assumptions, showing the
 advantage of the adaptive prior of MNTDP.
 Finally, EWC shows great transfer on
 $\mathcal{S}^{\mbox{\small{-}}}, \mathcal{S}^{\mbox{\tiny{in}}}, \mathcal{S}^{\mbox{\tiny{out}}}$,
 which probe the ability of the model to retain information. However,
 it fails at $\mathcal{S}^{\mbox{\small{+}}}, \mathcal{S}^{\mbox{\tiny{pl}}}$ that require
 additional capacity allocation  to learn a new task. Fig.~\ref{fig:radar_chart_first_page} gives a holistic
view by	reporting the normalized performance across all these 	dimensions.



\begin{figure}
\begin{floatrow}
\capbtabbox{%
   \begin{tabular}{p{1.6cm}p{.7cm}p{.7cm}p{.6cm}p{.6cm}}
\toprule
{} &  \hspace{-4mm}\mbox{$<\mathcal{A}>$} &  \hspace{-4mm}\mbox{$<\mathcal{F}>$} &  \hspace{-4mm}  Mem. & \mbox{\hspace{-3mm}PFLOPs} \\
\midrule

Independent      &  0.57  &     0.0   &    243 &   4    \\
Finetune        &    0.20  &   -0.4   &      2  &   5     \\
New-head         &  0.43  &     0.0 &      3  &   6  \\
On. EWC$\dagger$       &  0.27   &  -0.3   &      7   &   4    \\
MNTDP-S         &   0.68   &     0.0   &  159  &   5   \\
\mbox{MNTDP-D}         &   0.75  &     0.0   &   102  &  26   \\
\hline
\mbox{MNTDP-D*}        &   0.75   &     0.0   &   1782   &  3  \\
HAT*$\dagger$            &  0.24  &   -0.1  &     32   &      $\approx$0  \\
\mbox{HAT*$\dagger$ (Wide)}     &   0.32   &     0.0   &    285  &   1  \\

\bottomrule
\end{tabular}

}{%
  \caption{\label{tab:res-long-100} \small Results on the long evaluation stream. * correspond to models using an Alexnet backbone. See Tab. \ref{tab:res-long-stream-full} for more baselines and error bars.}%
}
\ffigbox{%
 \hspace{-1.5em}\includegraphics[width=7cm]{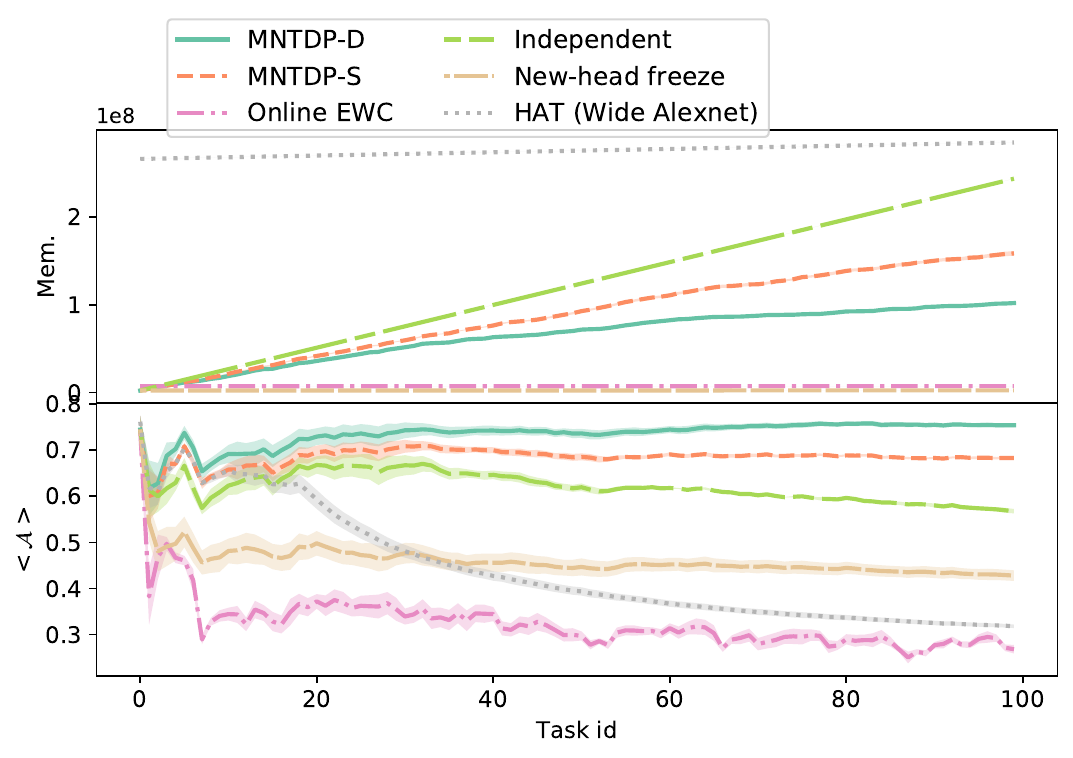}
  \vspace{-.2cm}
}{%
  \caption{\label{fig:long_stream} \small Evolution of $<\mathcal{A}>$ and Mem. on $\mathcal{S}^{\mbox{\tiny{long}}}$.}%
}
\end{floatrow}

  \vspace{-.2cm}
\end{figure}

 
We conclude by reporting results on
$\mathcal{S}^{\mbox{\tiny{long}}}$ composed of 100
tasks. Tab.~\ref{tab:res-long-100} reports the results of all the
approaches we could train without running into out-of-memory. MNTDP-D yields an absolute 18\% improvement over the baseline
\textit{independent model} while using less than half of its memory,
thanks to the discovery of many paths with shared modules.
Its actual runtime is close to \textit{independent model} because of
GPU parallel computing. 
To match the capacity of MNTDP, we scale HAT's backbone to the maximal
size that can fit in a Titan X GPU Memory (6.5x, wide version). The
wider architecture greatly increases inference time in later tasks (see also discussion on memory complexity at test time in Appendix~\ref{app:mem_complexity}),
while our modular approach uses the same backbone for every task and
yet it achieves better performance. Fig.~\ref{fig:long_stream} shows
the average accuracy up to the current task over time. MNTDP-D attains
the best performance while growing sublinearly in terms of memory
usage. Methods that do not evolve their architecture, like EWC, greatly suffer in
terms of average accuracy.

\paragraph{Ablation: } We first study the importance of the prior. Instead of
selecting the nearest neighbor path, we pick one path corresponding to one of the previous tasks at
random. In this case, $\mathcal{T}(\mathcal{S}^{\mbox{\small{-}}})$ decreases from $0$ to $-0.2$ and
$\mathcal{T}(\mathcal{S}^{out})$ goes from $0$ to $-0.3$. With a
random path, MNTDP learns not to share any module, demonstrating 
that it is indeed important to form a good prior over the search
space. Appendix~\ref{app:detailed_results} reports additional results
demonstrating how on small streams MNTDP is robust to the choice of $k$ in
the prior since we attain similar performance using $k=1$ and $k=all$, although only $k=1$ let us scale to $\mathcal{S}^{\mbox{\tiny{long}}}$.
Finally, we explore the robustness to the number of modules by
splitting each module in two, yielding a total of 10 modules per path, and by
merging adjacent modules yielding a total of 3 modules for the same
overall number of parameters. We find that
$\mathcal{T}(\mathcal{S}^{\mbox{\tiny{out}}})$ decreases from 0 to -0.1, with a 9\% decrease on $t_1^-$, when the number of modules decreases but stays the same when the number of modules increases, suggesting that the algorithm has to have a sufficient number of modules to flexibly grow.

\section{Conclusions} \label{sec:conclusions}
We introduced a new benchmark to enable a more comprehensive
evaluation of CL algorithms, not only in terms of average accuracy and forgetting but
also knowledge transfer and scaling. We have also
proposed a modular network that can gracefully scale thanks to an adaptive prior over the
search space of possible ways to connect modules.
Our experiments show that our approach yields a very desirable
trade-off between accuracy and compute/memory usage, suggesting that
modularization in a restricted search space is a promising avenue of investigation for continual
learning and knowledge transfer.



\bibliography{iclr2021_conference}
\bibliographystyle{iclr2021_conference}

\newpage
\appendix

\section{Datasets and Streams} \label{app:datasets}

\begin{figure}[!h]	
	\centering
	\begin{subfigure}[t]{.2\linewidth}
		\centering
		\includegraphics[width=.85\textwidth]{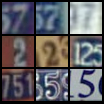}
		\caption{SVHN}\label{fig:ds-svhn}		
	\end{subfigure}
	\quad
	\begin{subfigure}[t]{0.2\textwidth}
		\centering
		\includegraphics[width=.8\textwidth]{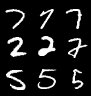}
		\caption{MNIST}\label{fig:ds-mnist}		
	\end{subfigure}
	\quad
	\begin{subfigure}[t]{0.2\textwidth}
		\centering
		\includegraphics[width=.80\textwidth]{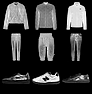}
		\caption{Fashion-MNIST}\label{fig:ds-fmnist}		
	\end{subfigure}
	\quad
	\begin{subfigure}[t]{0.2\textwidth}
		\centering
		\includegraphics[width=.85\textwidth]{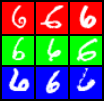}
		\caption{Rainbow-MNIST}\label{fig:ds-rmnist}		
	\end{subfigure}

	\caption{Some datasets used in the CTrL benchmark.}\label{fig:dataset_details}
\end{figure}

The CTrL benchmark is built using the standard datasets listed in
Tab.~\ref{tab:tasks-datasets}. Some examples from these datasets are
shown in Fig.~\ref{fig:dataset_details}. 
The composition of the different tasks
is given in Tab.~\ref{tab:stream-inter-reset} and an instance of the long stream is presented in Tab.
\ref{tab:long-stream-tasks-p1} and \ref{tab:long-stream-tasks-p2}.

The tasks in $\mathcal{S}^{-}$, $\mathcal{S}^{+}$,
$\mathcal{S}^{\mbox{\tiny{in}}}$, $\mathcal{S}^{\mbox{\tiny{out}}}$ and
$\mathcal{S}^{\mbox{\tiny{pl}}}$ are all 10-way classification tasks.
In $\mathcal{S}^{-}$, the first task has 4000 training examples while
the last one which is the same as the first task, has only 400.
The vice versa is true for $\mathcal{S}^{+}$ instead.
The last task of $\mathcal{S}^{\mbox{\tiny{in}}}$ is the same as the first task,
except that the background color of the MNIST digit is different.
The last task of $\mathcal{S}^{\mbox{\tiny{out}}}$ is the same as the first task,
except that label ids have been shuffled, therefore, if ``horse'' was
associated to label id $3$ in the first task, it is now associated to
label id $5$ in the last task.

        The $\mathcal{S}^{\mbox{\tiny{long}}}$ stream is composed of both \textit{large} and
        \textit{small} tasks that have $5000$ (or whatever is the
        maximum available) and $25$
        training examples, respectively. Each task is built by choosing one of the datasets at random,
and 5 categories at random in this dataset. During task 1-33, the fraction of small tasks is
50\%, this increases to 75\% for tasks 34-66, and to 100\% for tasks 67-100.
This is a challenging setting allowing to assess not only
 scalability, but also transfer ability and sample efficiency of a learner.                                      

 Scripts to rebuild the given streams and evaluate models will be released upon publication.

\begin{table}[h]
\begin{center}
\begin{tabular}{lllll}
\hline
Dataset     & no. classes & training & validation & testing \\ \hline
CIFAR-100   & 100         & 40000    & 10000      & 10000   \\
CIFAR-10    & 10          & 40000    & 10000      & 10000   \\
D. Textures & 47          & 1880     & 1880       & 1880    \\
SVHN        & 10          & 47217    & 26040      & 26032   \\
MNIST       &  10      & 50000  & 10000 & 10000           \\
Fashion-MNIST       &  10      & 50000  & 10000 & 10000           \\
Rainbow-MNIST & 10 & 50000 &10000 & 10000 \\
\end{tabular}
\caption{\label{tab:tasks-datasets}\small Datasets used in the CTrL benchmark.}
\end{center}
\end{table}

\begin{table}[ht]
  \centering
    \small

    \begin{tabular}{llllllll}
    \hline
    \textbf{Stream}                                      &                           & \textbf{$T_1$} & \textbf{$T_2$} & \textbf{$T_3$} & \textbf{$T_4$} & \textbf{$T_5$} & \textbf{$T_6$} \\ \hline \hline
    \multirow{3}{*}{\textbf{$\mathcal{S}^-$}}            & \textbf{Datasets}         & Cifar-10       & MNIST          & DTD            & F-MNIST        & SVHN           & Cifar-10       \\
                                                         & \textbf{\# Train Samples} & 4000           & 400            & 400            & 400            & 400            & 400            \\
                                                         & \textbf{\# Val Samples}   & 2000           & 200            & 200            & 200            & 200            & 200            \\ \hline
    \multirow{3}{*}{\textbf{$\mathcal{S}^+$}}            & \textbf{Datasets}         & Cifar-10       & MNIST          & DTD            & F-MNIST        & SVHN           & Cifar-10       \\
                                                         & \textbf{\# Train Samples} & 400            & 400            & 400            & 400            & 400            & 4000           \\
                                                         & \textbf{\# Val Samples}   & 200            & 200            & 200            & 200            & 200            & 2000           \\ \hline
    \multirow{3}{*}{\textbf{$\mathcal{S}^{\mbox{in}}$}}  & \textbf{Datasets}         & R-MNIST        & Cifar-10       & DTD            & F-MNIST        & SVHN           & R-MNIST        \\
                                                         & \textbf{\# Train Samples} & 4000           & 400            & 400            & 400            & 400            & 50             \\
                                                         & \textbf{\# Val Samples}   & 2000           & 200            & 200            & 200            & 200            & 30             \\ \hline
    \multirow{3}{*}{\textbf{$\mathcal{S}^{\mbox{out}}$}} & \textbf{Datasets}         & Cifar-10       & MNIST          & DTD            & F-MNIST        & SVHN           & Cifar-10       \\
                                                         & \textbf{\# Train Samples} & 4000           & 400            & 400            & 400            & 400            & 400            \\
                                                         & \textbf{\# Val Samples}   & 2000           & 200            & 200            & 200            & 200            & 200            \\ \hline
    \multirow{3}{*}{\textbf{$\mathcal{S}^{\mbox{pl}}$}}  & \textbf{Datasets}         & MNIST          & DTD            & F-MNIST        & SVHN           & Cifar-10       &                \\
                                                         & \textbf{\# Train Samples} & 400            & 400            & 400            & 400            & 4000           &                \\
                                                         & \textbf{\# Val Samples}   & 200            & 200            & 200            & 200            & 2000           &                \\ \hline
    \end{tabular}

\caption{Details of the streams used to evaluate the transfer properties of the learner.  F-MNIST is Fashion-MNIST and R-MNIST is a variant of Rainbow-MNIST, using only different background colors and keepingthe original scale and rotation of the digits \label{tab:stream-inter-reset}}
\end{table}

\newpage
\begin{table}[ht!]
  \centering
  \small
    \begin{tabular}{llllll}
\toprule
{} &        Dataset &                                            Classes &  \# Train &  \# Val &  \# Test \\
\textbf{Task id} &                &                                                    &          &        &         \\
\midrule
\textbf{1      } &          mnist &                                    [6, 3, 7, 5, 0] &       25 &     15 &    4830 \\
\textbf{2      } &           svhn &                                    [2, 1, 9, 0, 7] &       25 &     15 &    5000 \\
\textbf{3      } &           svhn &                                    [2, 0, 6, 1, 5] &     5000 &   2500 &    5000 \\
\textbf{4      } &           svhn &                                    [1, 5, 0, 7, 4] &       25 &     15 &    5000 \\
\textbf{5      } &  fashion-mnist &  [T-shirt/top, Pullover, Trouser, Sandal, Sneaker] &       25 &     15 &    5000 \\
\textbf{6      } &  fashion-mnist &  [Shirt, Ankle boot, Sandal, Pullover, T-shirt/... &     5000 &   2500 &    5000 \\
\textbf{7      } &           svhn &                                    [3, 1, 7, 6, 9] &       25 &     15 &    5000 \\
\textbf{8      } &       cifar100 &      [spider, maple\textbackslash \_tree, tulip, leopard, lizard] &       25 &     15 &     500 \\
\textbf{9      } &        cifar10 &           [frog, automobile, airplane, cat, horse] &       25 &     15 &    5000 \\
\textbf{10     } &  fashion-mnist &    [Ankle boot, Bag, T-shirt/top, Shirt, Pullover] &       25 &     15 &    5000 \\
\textbf{11     } &          mnist &                                    [4, 8, 7, 6, 3] &     5000 &   2500 &    4914 \\
\textbf{12     } &        cifar10 &              [automobile, truck, dog, horse, deer] &     5000 &   2500 &    5000 \\
\textbf{13     } &       cifar100 &          [sea, forest, bear, chimpanzee, dinosaur] &       25 &     15 &     500 \\
\textbf{14     } &          mnist &                                    [3, 2, 9, 1, 7] &       25 &     15 &    5000 \\
\textbf{15     } &  fashion-mnist &           [Bag, Ankle boot, Trouser, Shirt, Dress] &       25 &     15 &    5000 \\
\textbf{16     } &        cifar10 &                 [frog, cat, horse, airplane, deer] &       25 &     15 &    5000 \\
\textbf{17     } &        cifar10 &              [bird, frog, ship, truck, automobile] &     5000 &   2500 &    5000 \\
\textbf{18     } &           svhn &                                    [0, 4, 7, 5, 6] &     5000 &   2500 &    5000 \\
\textbf{19     } &          mnist &                                    [6, 5, 9, 4, 8] &     5000 &   2500 &    4806 \\
\textbf{20     } &          mnist &                                    [8, 5, 6, 4, 9] &     5000 &   2500 &    4806 \\
\textbf{21     } &       cifar100 &         [sea, pear, house, spider, aquarium\textbackslash \_fish] &       25 &     15 &     500 \\
\textbf{22     } &       cifar100 &            [kangaroo, ray, tank, crocodile, table] &     2250 &    250 &     500 \\
\textbf{23     } &       cifar100 &                  [trout, rose, pear, lizard, baby] &       25 &     15 &     500 \\
\textbf{24     } &           svhn &                                    [3, 2, 8, 1, 5] &     5000 &   2500 &    5000 \\
\textbf{25     } &       cifar100 &           [skyscraper, bear, rocket, tank, spider] &       25 &     15 &     500 \\
\textbf{26     } &       cifar100 &     [telephone, porcupine, flatfish, plate, shrew] &     2250 &    250 &     500 \\
\textbf{27     } &       cifar100 &         [lawn\textbackslash \_mower, crocodile, tiger, bed, bear] &       25 &     15 &     500 \\
\textbf{28     } &           svhn &                                    [3, 7, 1, 5, 6] &       25 &     15 &    5000 \\
\textbf{29     } &  fashion-mnist &      [Ankle boot, Sneaker, T-shirt/top, Coat, Bag] &     5000 &   2500 &    5000 \\
\textbf{30     } &          mnist &                                    [6, 9, 0, 3, 7] &     5000 &   2500 &    4938 \\
\textbf{31     } &        cifar10 &               [automobile, truck, deer, bird, dog] &       25 &     15 &    5000 \\
\textbf{32     } &        cifar10 &            [dog, airplane, frog, deer, automobile] &     5000 &   2500 &    5000 \\
\textbf{33     } &           svhn &                                    [1, 9, 5, 3, 6] &     5000 &   2500 &    5000 \\
\textbf{34     } &       cifar100 &  [whale, orange, chimpanzee, poppy, sweet\textbackslash \_pepper] &       25 &     15 &     500 \\
\textbf{35     } &       cifar100 &               [worm, camel, bus, keyboard, spider] &       25 &     15 &     500 \\
\textbf{36     } &  fashion-mnist &      [T-shirt/top, Coat, Ankle boot, Shirt, Dress] &       25 &     15 &    5000 \\
\textbf{37     } &        cifar10 &                      [dog, deer, ship, truck, cat] &       25 &     15 &    5000 \\
\textbf{38     } &        cifar10 &                   [cat, dog, airplane, ship, deer] &     5000 &   2500 &    5000 \\
\textbf{39     } &           svhn &                                    [7, 6, 4, 2, 9] &       25 &     15 &    5000 \\
\textbf{40     } &          mnist &                                    [9, 7, 1, 3, 2] &       25 &     15 &    5000 \\
\textbf{41     } &       cifar100 &        [mushroom, butterfly, bed, boy, motorcycle] &       25 &     15 &     500 \\
\textbf{42     } &  fashion-mnist &        [Shirt, Pullover, Bag, Sandal, T-shirt/top] &       25 &     15 &    5000 \\
\textbf{43     } &       cifar100 &          [rabbit, bear, aquarium\textbackslash \_fish, bee, bowl] &       25 &     15 &     500 \\
\textbf{44     } &  fashion-mnist &       [Coat, T-shirt/top, Pullover, Shirt, Sandal] &       25 &     15 &    5000 \\
\textbf{45     } &  fashion-mnist &             [Pullover, Dress, Coat, Shirt, Sandal] &       25 &     15 &    5000 \\
\textbf{46     } &          mnist &                                    [3, 9, 7, 6, 4] &       25 &     15 &    4940 \\
\textbf{47     } &        cifar10 &                [deer, bird, dog, automobile, frog] &       25 &     15 &    5000 \\
\textbf{48     } &           svhn &                                    [8, 7, 1, 0, 4] &       25 &     15 &    5000 \\
\textbf{49     } &       cifar100 &      [forest, skunk, poppy, bridge, sweet\textbackslash \_pepper] &     2250 &    250 &     500 \\
\textbf{50     } &       cifar100 &   [caterpillar, can, motorcycle, rabbit, wardrobe] &       25 &     15 &     500 \\
\bottomrule
\end{tabular}

    \caption{\label{tab:long-stream-tasks-p1}\small Details of the tasks used
      in $\mathcal{S}^{\mbox{\tiny{long}}}$, part 1.}
\end{table}
\newpage
\begin{table}[ht!]
  \centering
  \small
    \begin{tabular}{llllll}
\toprule
{} &        Dataset &                                           Classes &  \# Train &  \# Val &  \# Test \\
\textbf{Task id} &                &                                                   &          &        &         \\
\midrule
\textbf{51     } &       cifar100 &      [trout, mountain, kangaroo, pine\textbackslash \_tree, bee] &       25 &     15 &     500 \\
\textbf{52     } &       cifar100 &           [clock, fox, castle, bus, willow\textbackslash \_tree] &       25 &     15 &     500 \\
\textbf{53     } &        cifar10 &                 [cat, airplane, dog, ship, truck] &       25 &     15 &    5000 \\
\textbf{54     } &          mnist &                                   [9, 7, 8, 1, 5] &       25 &     15 &    4866 \\
\textbf{55     } &  fashion-mnist &          [Bag, T-shirt/top, Sandal, Shirt, Dress] &       25 &     15 &    5000 \\
\textbf{56     } &  fashion-mnist &      [Sneaker, Ankle boot, Coat, Sandal, Trouser] &     5000 &   2500 &    5000 \\
\textbf{57     } &          mnist &                                   [1, 4, 3, 9, 7] &       25 &     15 &    4982 \\
\textbf{58     } &        cifar10 &              [truck, automobile, frog, ship, dog] &       25 &     15 &    5000 \\
\textbf{59     } &          mnist &                                   [7, 2, 8, 5, 4] &       25 &     15 &    4848 \\
\textbf{60     } &          mnist &                                   [2, 8, 9, 1, 7] &     5000 &   2500 &    4974 \\
\textbf{61     } &           svhn &                                   [9, 5, 1, 8, 6] &       25 &     15 &    5000 \\
\textbf{62     } &          mnist &                                   [1, 8, 7, 4, 5] &       25 &     15 &    4848 \\
\textbf{63     } &        cifar10 &          [truck, dog, bird, automobile, airplane] &       25 &     15 &    5000 \\
\textbf{64     } &          mnist &                                   [8, 4, 3, 7, 6] &       25 &     15 &    4914 \\
\textbf{65     } &           svhn &                                   [3, 5, 7, 2, 1] &       25 &     15 &    5000 \\
\textbf{66     } &       cifar100 &                  [otter, camel, bee, road, poppy] &       25 &     15 &     500 \\
\textbf{67     } &           svhn &                                   [4, 2, 1, 8, 7] &       25 &     15 &    5000 \\
\textbf{68     } &          mnist &                                   [3, 7, 6, 8, 9] &       25 &     15 &    4932 \\
\textbf{69     } &  fashion-mnist &       [Pullover, Sneaker, Trouser, Dress, Sandal] &       25 &     15 &    5000 \\
\textbf{70     } &           svhn &                                   [5, 0, 7, 2, 3] &       25 &     15 &    5000 \\
\textbf{71     } &           svhn &                                   [9, 6, 2, 4, 8] &       25 &     15 &    5000 \\
\textbf{72     } &          mnist &                                   [7, 1, 2, 0, 6] &       25 &     15 &    4938 \\
\textbf{73     } &        cifar10 &            [dog, automobile, ship, airplane, cat] &       25 &     15 &    5000 \\
\textbf{74     } &          mnist &                                   [0, 7, 6, 2, 4] &       25 &     15 &    4920 \\
\textbf{75     } &        cifar10 &                 [bird, deer, airplane, dog, ship] &       25 &     15 &    5000 \\
\textbf{76     } &       cifar100 &  [mountain, bicycle, caterpillar, spider, possum] &       25 &     15 &     500 \\
\textbf{77     } &           svhn &                                   [8, 3, 4, 0, 6] &       25 &     15 &    5000 \\
\textbf{78     } &           svhn &                                   [1, 5, 9, 0, 8] &       25 &     15 &    5000 \\
\textbf{79     } &       cifar100 &        [can, dolphin, house, pickup\textbackslash \_truck, crab] &       25 &     15 &     500 \\
\textbf{80     } &       cifar100 &  [squirrel, possum, crocodile, mountain, hamster] &       25 &     15 &     500 \\
\textbf{81     } &          mnist &                                   [7, 0, 1, 6, 2] &       25 &     15 &    4938 \\
\textbf{82     } &  fashion-mnist &     [T-shirt/top, Dress, Trouser, Shirt, Sneaker] &       25 &     15 &    5000 \\
\textbf{83     } &        cifar10 &            [cat, frog, automobile, dog, airplane] &       25 &     15 &    5000 \\
\textbf{84     } &        cifar10 &               [automobile, cat, dog, ship, horse] &       25 &     15 &    5000 \\
\textbf{85     } &       cifar100 &              [cup, otter, orchid, kangaroo, rose] &       25 &     15 &     500 \\
\textbf{86     } &          mnist &                                   [1, 5, 7, 2, 9] &       25 &     15 &    4892 \\
\textbf{87     } &           svhn &                                   [6, 5, 3, 2, 7] &       25 &     15 &    5000 \\
\textbf{88     } &        cifar10 &                      [dog, deer, cat, frog, bird] &       25 &     15 &    5000 \\
\textbf{89     } &          mnist &                                   [6, 2, 5, 9, 4] &       25 &     15 &    4832 \\
\textbf{90     } &       cifar100 &                 [pear, rocket, sea, road, orange] &       25 &     15 &     500 \\
\textbf{91     } &           svhn &                                   [0, 8, 4, 6, 1] &       25 &     15 &    5000 \\
\textbf{92     } &        cifar10 &                   [truck, horse, ship, deer, dog] &       25 &     15 &    5000 \\
\textbf{93     } &          mnist &                                   [5, 8, 6, 4, 3] &       25 &     15 &    4806 \\
\textbf{94     } &           svhn &                                   [2, 6, 3, 4, 1] &       25 &     15 &    5000 \\
\textbf{95     } &  fashion-mnist &       [Bag, Trouser, Sneaker, Ankle boot, Sandal] &       25 &     15 &    5000 \\
\textbf{96     } &           svhn &                                   [7, 9, 1, 5, 8] &       25 &     15 &    5000 \\
\textbf{97     } &       cifar100 &           [lamp, otter, skyscraper, sea, raccoon] &       25 &     15 &     500 \\
\textbf{98     } &       cifar100 &                [clock, flatfish, snake, can, man] &       25 &     15 &     500 \\
\textbf{99     } &           svhn &                                   [6, 3, 0, 8, 7] &       25 &     15 &    5000 \\
\textbf{100    } &  fashion-mnist &            [Shirt, Coat, Dress, Sandal, Pullover] &       25 &     15 &    5000 \\
\bottomrule
\end{tabular}

    \caption{\label{tab:long-stream-tasks-p2}\small Details of the task in
      $\mathcal{S}^{\mbox{\tiny{long}}}$, part 2.}
\end{table}
\clearpage
\newpage




\section{Learning Algorithm}

We provide a more complete description of the two learning algorithms
for MNTDP-S and MNTDP-D. In the deterministic case, all the
architectures (given how we restrict the search space, we have $7$ of them) are trained over the training set, and the best path is
retained based on its score on the validation set. To avoid overfitting when using MNDTDP-S, we split the training set into two halves
$\mathcal{D}^{t}_1$ and $\mathcal{D}^{t}_2$. The first part is used to
update the module parameters $\theta$, while the second is used to update the
parameters in the distribution over paths, $\Gamma$. If both sets of parameters were trained on the same dataset, $\Gamma$ would favor paths prone to overfitting since they will results in a large decrease of its training loss and therefore a larger reward. When they are trained on different sets, $\Gamma$ has to select paths with a reasonable amount of free parameters, allowing $\theta$ to learn and generalize enough to decrease the loss on both training sets.  
Then, the most promising architecture
 is selected based on $\arg\max \Gamma$, and fine-tuned over the whole
 $\mathcal{D}^t$.
 In this case, the validation set is used only for hyper-parameters selection.

\begin{minipage}{0.46\textwidth}
\begin{algorithm}[H]
\caption{MNTDP-S algorithm. 
\label{algo:MNTDP-S}}
\SetAlgoLined
\scriptsize
\nl \KwData{Dataset of task $t$: $\mathcal{D}^t$.\;}
\nl {\bf Past predictors:} $f(x, j | \mathcal{S})$ for $j=1,\dots,t-1$ \;
\nl {\bf Find closest task:} $j^* = \arg \max_j \mathbb{E}_{(x,y) \sim \mathcal{D}^t} [                                                            
        \Delta(\mbox{NN}(f(x, j |
        \mathcal{S})), y)]$, where NN is the 5-nearest neighbor
classifier in feature space\;
\nl {\bf Define search space:} Take path corresponding to predictor of
task $j^*$ and add a new randomly initialized module at every
layer. $\Gamma$: distribution over paths; $\pi_i$: $i$-th path\;
\nl {\bf Split train set in two halves:} $\mathcal{D}^{t}_1$ and
$\mathcal{D}^{t}_2$\;
\nl \While{loss in eq.~ \ref{eq:loss_general} has not converged} {
  \nl get sample $(x,y) \sim \mathcal{D}^t[iteration \textrm{ mod } 2]$\;
  \nl sample path $\pi_k \sim \Gamma$\;
  \nl \eIf{odd iteration \textbf{or} $\max_{\Gamma} > 0.99$}{
    \nl forward/backward and update $\theta(\pi_k)$ (only newly added modules)
    
  }{
    \nl forward/backward and update $\Gamma$\;
  }
}
\nl Let $i^*$ be the path with largest values in $\Gamma$, then set $f(x, t |
\mathcal{S} \cup t )$ to $\pi_{i^*}$.

\end{algorithm}
\end{minipage}
\hfill
\begin{minipage}{0.46\textwidth}
    \begin{algorithm}[H]
      \SetAlgoLined
      \scriptsize
      \nl \KwData{Dataset of task $t$: $\mathcal{D}^t$.\;}
      \nl {\bf Past predictors:} $f(x, j | \mathcal{S})$ for
      $j=1,\dots,t-1$ \;
      \nl {\bf Find closest task:} $j^* = \arg \max_j
      \mathbb{E}_{(x,y) \sim \mathcal{D}^t} [
        \Delta(\mbox{NN}(f(x, j |
        \mathcal{S})), y)]$, where NN is 5-nearest neighbor classifier
      in feature space\;
      \nl {\bf Define search space:} Take path corresponding to
      predictor of task $j^*$ and add a new randomly initialized
      module at every layer. $\pi_i$: $i$-th path, $i = 1,\dots,N$
      where $N$ is the total number of paths\;
      \nl \For{$i = 1, \dots, N$}{
        \nl \While{loss in eq.~\ref{eq:loss_general} has not
          converged} {
          \nl get sample $(x,y) \sim \mathcal{D}^t$\;
          \nl forward/backward, update parameters $\theta(\pi_i)$
          (only newly added modules)
        }
        \nl compute accuracy $A_i$ on validation set.
      }
      \nl Let $i^* = \arg \max_i A_i$, then set $f(x, t |                                                                                                
\mathcal{S} \cup t )$ to $\pi_{i^*}$.
      \caption{MNTDP-D algorithm.
        \label{algo:MNTDP-D}}
    \end{algorithm}
  \end{minipage}




\section{Architecture of the Models and Hyper-Parameters} \label{app:models}
The model used on Permuted-MNIST is presented in table \ref{tab:fc-arch-details}. On all other tasks, the base architecture for all baselines and MNTDP is the same CNN, a
ResNet convolutional neural network as reported in
Table~\ref{tab:resnet-arch-details} or AlexNet as reported in Table~\ref{tab:alexnet-arch-details}.
The table also shows how layers are grouped into modules for MNTDP and PNN.

\begin{table}[h]
\centering
\begin{tabular}{cccc}
\hline
\textbf{Block} & \textbf{\# layers} & \textbf{\#params} & \textbf{\# hidden units} \\ \hline
1              & 1                  & 785000                & 1000                  \\
2              & 1                  & 1001000               & 1000               \\
3              & 1                  & 10010                  & num. classes           \\ \hline
\end{tabular}
\caption{Permuted MNIST Model}
\label{tab:fc-arch-details}

\end{table}

\begin{table}[h]
\small
  \centering
\begin{tabular}{cccc}
\hline
\textbf{Block} & \textbf{\# layers} & \textbf{\#params} & \textbf{\# out channels} \\ \hline
1              & 1                  & 1856              & 64                       \\
2              & 4                  & 147968            & 64                       \\
3              & 4                  & 152192            & 64                       \\
4              & 4                  & 152192            & 64                       \\
5              & 2                  & 78208             & 64                      \\
6              & 2                  & 73984             & 64                      \\
7              & 1                  & 650               & num. classes                        \\ \hline
\end{tabular}
\caption{Resnet architecture used throughout our experiments.}
\label{tab:resnet-arch-details}
\end{table}
	
\begin{table}[h]
\small
  \centering
\begin{tabular}{cccc}
\hline
\textbf{Block} & \textbf{\# layers} & \textbf{\#params} & \textbf{\# out channels/hidden units} \\ \hline
1              & 1                  & 3136              & 64                       \\
2              & 1                  & 73856            & 128                       \\
3              & 1                  & 131328            & 256                       \\
4              & 1                  & 2099200            & 2048                       \\
5              & 1                  & 4196352             & 2048                      \\
7              & 1                  & 10245               & num. classes                        \\ \hline
\end{tabular}
\caption{Details of the Alexnet architecture used in \citet{HAT} and in our experiments.}
\label{tab:alexnet-arch-details}
\end{table}


As presented in
section~\ref{sec:exp_modeling}, the stream can be visited only once,
preventing stream-level hyper-parameters tuning. Exceptions are made
for HAT~\citep{HAT} because we have been using authors' implementation, and for EWC and Online-EWC since these approaches fail in the proposed
setting. The constraint strength hyper-parameter $\lambda$ must be
tuned at the stream level since a task-level tuning of $\lambda$
results in little or no constraint at all, leading to severe catastrophic
forgetting.
The stream-level hyper-parameters optimization considers 9 values for $\lambda$ \{1, 5, 10, 50, 100, 500, $10^3$, $5\times10^3$, $10^4$\}.
Note that this gives an unfair advantage to EWC,  Online-EWC and HAT, as all
other methods including MNTDP use task-level cross-validation as
described in \textsection\ref{sec:exp_modeling}.

For all methods and experiments, we use the Adam optimizer~\citep{DBLP:journals/corr/KingmaB14} with $\beta_1 = 0.9$, $\beta_2 = 0.999$ and $\epsilon = 10^{-8}$.

For each task and each baseline, two learning rates \{$10^{-2}$,
$10^{-3}$\} and 3 weight decay strengths \{$0$, $10^{-5}$, $10^{-4}$\}
are considered. Early stopping is performed on each task to identify
the best step to stop training. When the current task validation
accuracy stops increasing for 300 iterations, we restore the learner
to its state after the best iteration and stop training on the current
task. 

For MNTDP-S, we consider two additional learning rates for the
$\Gamma$ optimization \{$10^{-2}$, $10^{-3}$\}. An entropy
regularization term on $\Gamma$ is added to the loss to encourage
exploration, preventing an early convergence towards a sub-optimal
path. The weight for this regularization term is set to 1 throughout our experiments 

Finally, since small tasks in $\mathcal{S}^{\mbox{\tiny{long}}}$ have very
few examples in the validation sets, we use test-time augmentation to prevent
overfitting during the grid search. For each validation sample, we add
four augmented copies following the same data augmentation procedure
used during training.

\section{Memory complexity of the Models}\label{app:mem_complexity}
In the main paper we report the overall memory consumption by the end of training. However, modular architectures like MNTDP use only a sparse subset of modules for a given task. In this section, we report the memory complexity both at training and test time. Table~\ref{tab:mem_complex} shows that all methods but DEN and RCL are as fast to evaluate at test time as an independent network because they all use a single path.
\begin{table}[h]
\begin{tabular}{lll}
\hline
                        & \textbf{Train} & \textbf{Test} \\ \hline
Inde, fintune, new-head & $N$            & $N$           \\
EWC                     & $N+2TN$        & $N$           \\
ER                      & $N+rT$         & $N$           \\
MNTDP-D                 & $kbN$          & $N$           \\
MNTDP-S                 & $N+2kN$        & $N$           \\
HAT                     & $N$            & $N$           \\
Wide HAT                & $SN$           & $SN$          \\
DEN                     & $N+pT$         & $N+pT$          \\
RCL                     & $N+pT$         & $N+pT$        \\
Lean to Grow            & $N+2Tp$        & $N$           \\ \hline
\end{tabular}
\caption{Memory complexity of the different baselines at train time and at test time, where $N$ is the size of the backbone, $T$ the number of tasks, $r$ the size of the memory buffer per task, $k$ the number of source columns used by MNTDP, $b$ the number of blocks in the backbone, $S$ the scale factor used for wide-HAT and $p$ the average number of new parameters introduced per task. Note that while Wide HAT and MNTDP-D are using a similar amount of memory on CTrL (Table \ref{tab:res-all-metrics}), the inference model used by MNTDP on each task only uses the memory of the narrow backbone, resulting in more than 6 times smaller inference models.}
\label{tab:mem_complex}
\end{table}

\section{Additional Results} \label{app:detailed_results}
\subsection{metric}

\paragraph{LCA} In addition to the metrics used in the main paper, we report the
area under the learning curve after $\beta$  optimization
steps~\citep{DBLP:conf/iclr/ChaudhryRRE19} as a metric to assess
learning speed:
$$\textrm{LCA}@\beta = \frac{1}{T}\sum\limits_{t=1}^T  \left[ \frac{1}{\beta+1}\sum\limits_{b=0}^\beta A_{b,t}(f) \right]$$
Where $A_{b,t}(f)$ corresponds to the test set accuracy on task $t$ of learner $f$ after having seen $b$ batches from $t$

\newpage

\subsection{Classical Streams} \label{app:classic_streams}
Table \ref{tab:res-mnist-perm}  and \ref{tab:res-cifar100-split}
report performance across all axes of evaluation on the standard
Permuted MNIST and Split CIFAR 100 streams.


    \begin{table}[h]
        \centering

\begin{tabular}{lccccc}
\toprule
{} &  $<\mathcal{A}>$ &  $<\mathcal{F}>$ &   Mem. &  FLOPs &  LCA@5 \\
\textbf{Model             } &                  &                  &        &        &        \\
\midrule
\textbf{Independent       } &             0.98 &             0.00 &   71.8 &    9.0 &   0.43 \\
\textbf{Finetune          } &             0.49 &            -0.49 &    7.2 &   12.0 &   0.13 \\
\textbf{New-head freeze   } &             0.89 &             0.00 &    7.5 &   19.0 &   0.14 \\
\textbf{New-head finetune } &             0.55 &            -0.43 &    7.5 &   15.0 &   0.14 \\
\textbf{New-leg freeze    } &             0.98 &             0.00 &   35.4 &   10.0 &   0.43 \\
\textbf{New-leg finetune  } &             0.89 &            -0.09 &   35.4 &   14.0 &   0.19 \\
\textbf{EWC $\dagger$               } &             0.94 &            -0.03 &  143.7 &   18.0 &   0.13 \\
\textbf{Online EWC  $\dagger$      } &             0.95 &            -0.01 &   21.6 &   14.0 &   0.16 \\
\textbf{ER (Reservoir) $\dagger$   } &             0.69 &            -0.29 &   15.0 &   12.0 &   0.29 \\
\textbf{ER                } &             0.90 &            -0.08 &   15.0 &   15.0 &   0.29 \\
\textbf{PNN               } &             0.98 &             0.00 &  253.8 &  123.0 &   0.15 \\
\textbf{MNTDP-S           } &             0.97 &             0.00 &   71.8 &   21.0 &   0.22 \\
\textbf{MNTDP-S (k=all)   } &             0.98 &             0.00 &   71.8 &   24.0 &   0.19 \\
\textbf{MNTDP-D           } &             0.98 &             0.00 &   71.8 &   56.0 &   0.37 \\
\textbf{HAT$\dagger$ } &             0.95 &             0.00 &    7.6 &   25.0 &   0.11 \\
\textbf{HAT (Wide)$\dagger$} &             0.97 &             0.00 &   78.5 &  209.0 &   0.13 \\
\textbf{DEN $\dagger$     } &             0.95 &             0.00 &      8.1     &   - &   - \\
\textbf{RCL $\dagger$     } &             0.96 &             0.00 &      8.5     &   148.2 &   - \\
\bottomrule
\end{tabular}

        \caption{\label{tab:res-mnist-perm} \small Results on the standard
          permuted-MNIST stream. In this stream, each of the 10 tasks
          corresponds to a random permutation of the input pixels of MNIST
          digits. For DEN and RCL, since we are using the authors' implementations, we do not have access to the LCA measure. $\dagger$ correspond to models using stream-level cross-validation (see Section 5.2).}
    \end{table}

\begin{table}[h]
    \centering

\begin{tabular}{lccccc}
\toprule
{} &  $<\mathcal{A}>$ &  $<\mathcal{F}>$ &   Mem. &   FLOPs &  LCA@5 \\
\textbf{Model             } &                  &                  &        &         &        \\
\midrule
\textbf{Independent       } &             0.82 &             0.00 &   24.4 &  1327.0 &   0.11 \\
\textbf{Finetune          } &             0.18 &            -0.56 &    2.4 &   886.0 &   0.15 \\
\textbf{New-head freeze   } &             0.57 &             0.00 &    2.5 &   740.0 &   0.21 \\
\textbf{New-head finetune } &             0.21 &            -0.52 &    2.5 &   687.0 &   0.15 \\
\textbf{New-leg freeze    } &             0.50 &             0.00 &    2.5 &  1759.0 &   0.12 \\
\textbf{New-leg finetune  } &             0.18 &            -0.42 &    2.5 &  1115.0 &   0.11 \\
\textbf{EWC      $\dagger$         } &             0.55 &             0.01 &   51.0 &  1151.0 &   0.11 \\
\textbf{Online EWC  $\dagger$      } &             0.54 &            -0.02 &    7.3 &  1053.0 &   0.12 \\
\textbf{ER (Reservoir) $\dagger$   } &             0.32 &            -0.50 &   20.9 &  1742.0 &   0.17 \\
\textbf{ER                } &             0.66 &            -0.15 &   20.9 &  1524.0 &   0.17 \\
\textbf{PNN               } &             0.78 &             0.00 &  133.8 &  9889.0 &   0.15 \\
\textbf{MNTDP-S           } &             0.75 &             0.00 &   24.4 &  1295.0 &   0.11 \\
\textbf{MNTDP-S (k=all)   } &             0.75 &             0.00 &   24.4 &  1323.0 &   0.11 \\
\textbf{MNTDP-D           } &             0.83 &             0.00 &   24.3 &  6168.0 &   0.12 \\
\textbf{MNTDP-D* } &             0.81 &             0.00 &  260.9 &   488.0 &   0.17 \\
\textbf{HAT*$\dagger$     } &             0.74 &            -0.01 &   27.0 &   175.0 &   0.11 \\
\textbf{HAT (Wide)*$\dagger$} &             0.79 &             0.00 &  269.3 &  1830.0 &   0.11 \\
\bottomrule
\end{tabular}

    \caption{\label{tab:res-cifar100-split} \small Results on the standard
      Split Cifar 100 stream. Each task is composed of 10 new classes.* correspond to models using an Alexnet backbone, $\dagger$ to models using stream-level cross-validation (see Section 5.2).}
\end{table}
\clearpage
\newpage
\subsection{The CTrL benchmark}
\begin{figure}[h]
        \centering
    \includegraphics[width=\textwidth]{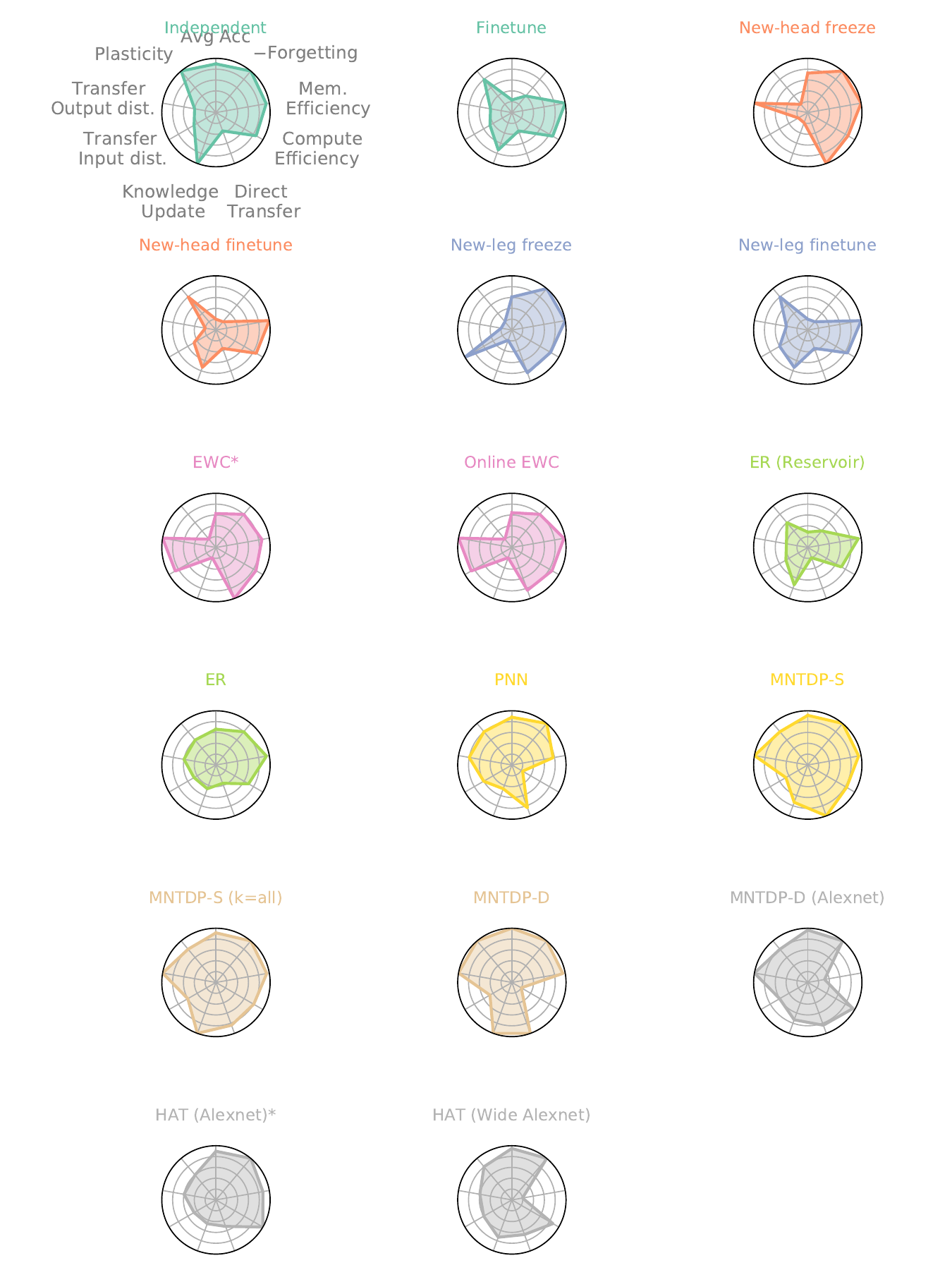}
        \caption{\small Comparison of the global performance of all baselines
          on the CTrL Benchmark. MNTDP-D is the
    most efficient method on multiple of the dimensions, but it
    requires more computation than MNTDP-S.
        }\label{fig:radar-all}
\end{figure}

Figure~\ref{fig:radar_chart_first_page} and \ref{fig:radar-all} provide radar plots of the models
evaluated on the CTrL benchmark. The companion tables of these
plots are in Tab.~\ref{tab:res-remember-full},
\ref{tab:res-reset-full}, \ref{tab:res-leg-full},
\ref{tab:res-head-full} and \ref{tab:res-plasticity-full}.

\clearpage
\newpage
\subsection{Discovered paths}\label{app:discovered_paths}
\begin{figure}[h]
        \centering
    \includegraphics[width=\textwidth]{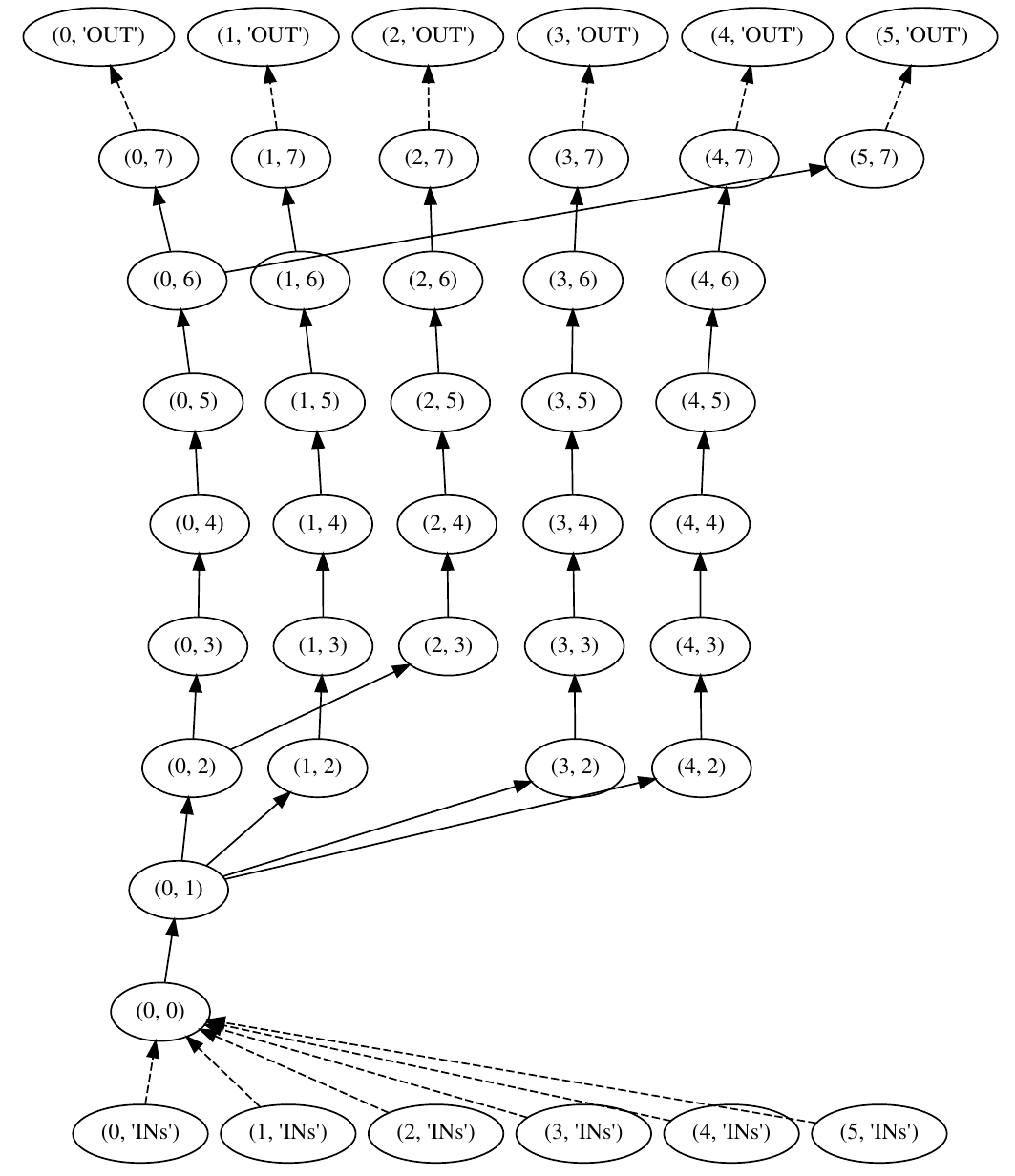}
        \caption{\small Global graph of paths discovered by MNDTP-D on the $\mathcal{T}(\mathcal{S}^{\mbox{\tiny{out}}})$ Stream. "INs" (resp. "OUT") nodes are the input (resp. output) of the path for each task. Solid edges correspond to parameterized modules while dashed edges are only used to show which block is selected for each task and don't apply any transformation. We observe that a significant amount of new parameters are introduced for tasks 2, 3, 4 and 5, which are very different from the first task. The model is however able to correctly identify that the last task is very similar to the first one, resulting in very large reuse of past modules and only introducing a new classification layer to adapt to the new task. }
\end{figure}

\begin{figure}[h]
        \centering
    \includegraphics[width=\textwidth]{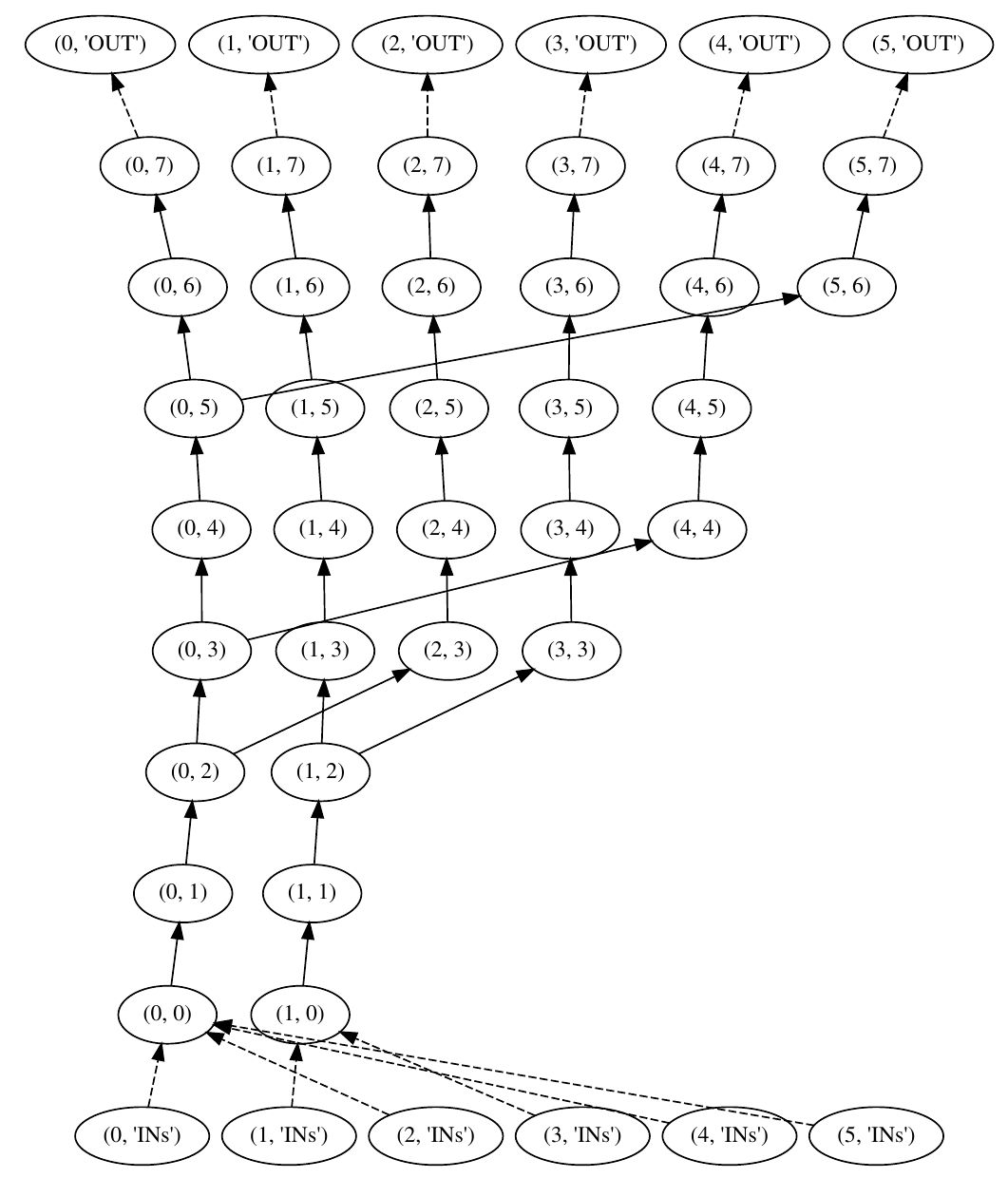}
        \caption{\small Global graph of paths discovered by MNDTP-S on the $\mathcal{T}(\mathcal{S}^{\mbox{\tiny{out}}})$ Stream. When facing the last task, the model correctly identified that modules from the first task should be reused, ultimately introducing 2 new modules to solve it. }
\end{figure}
\clearpage
\newpage

\subsection{Stream $\mathcal{S}^{\mbox{\small{-}}}$}
\begin{table}[h]
    \centering
    \begin{tabular}{lp{.8cm}p{.9cm}p{.8cm}p{.8cm}p{1cm}p{1cm}p{.8cm}p{.8cm}p{.8cm}}
\toprule
{} &  \mbox{Acc $T_1$} &  \mbox{Acc $T_1'$} &  $\Delta_{T_1, T_1'}$ &  $\mathcal{T}(\mathcal{S}^-)$ &  \mbox{$<\mathcal{A}>$} &  \mbox{$<\mathcal{F}>$} &   Mem. &   FLOPs &  LCA@5 \\
\textbf{Model             } &            &             &                       &                    &                  &                  &        &         &        \\
\midrule
\textbf{Independent       } &       0.72 &        0.36 &                 -0.35 &               0.00 &             0.56 &             0.00 &   14.6 &   292.0 &   0.10 \\
\textbf{Finetune          } &       0.72 &        0.37 &                 -0.34 &               0.01 &             0.18 &            -0.33 &    2.4 &   308.0 &   0.10 \\
\textbf{New-head freeze   } &       0.72 &        0.71 &                 -0.01 &               0.35 &             0.54 &             0.00 &    2.5 &   262.0 &   0.19 \\
\textbf{New-head finetune } &       0.72 &        0.35 &                 -0.36 &              -0.01 &             0.15 &            -0.39 &    2.5 &   296.0 &   0.10 \\
\textbf{New-leg freeze    } &       0.72 &        0.65 &                 -0.07 &               0.29 &             0.47 &             0.00 &    2.5 &   362.0 &   0.11 \\
\textbf{New-leg finetune  } &       0.72 &        0.33 &                 -0.38 &              -0.03 &             0.13 &            -0.41 &    2.5 &   333.0 &   0.10 \\
\textbf{EWC $\dagger$               } &       0.72 &        0.71 &                 -0.01 &               0.35 &             0.52 &            -0.02 &   31.5 &   344.0 &   0.13 \\
\textbf{Online EWC  $\dagger$      } &       0.72 &        0.69 &                 -0.03 &               0.33 &             0.54 &            -0.01 &    7.3 &   309.0 &   0.11 \\
\textbf{ER (Reservoir)$\dagger$    } &       0.72 &        0.30 &                 -0.41 &              -0.06 &             0.20 &            -0.32 &   13.5 &   646.0 &   0.12 \\
\textbf{ER                } &       0.72 &        0.36 &                 -0.36 &               0.00 &             0.41 &            -0.13 &   13.5 &   551.0 &   0.11 \\
\textbf{PNN               } &       0.72 &        0.65 &                 -0.06 &               0.29 &             0.62 &             0.00 &   51.1 &  1099.0 &   0.13 \\
\textbf{MNTDP-S           } &       0.72 &        0.71 &                  0.00 &               0.35 &             0.63 &             0.00 &   11.0 &   310.0 &   0.10 \\
\textbf{MNTDP-S (k=all)   } &       0.72 &        0.63 &                 -0.09 &               0.27 &             0.61 &             0.00 &   10.7 &   341.0 &   0.10 \\
\textbf{MNTDP-D           } &       0.72 &        0.72 &                  0.00 &               0.36 &             0.67 &             0.00 &    9.2 &  1876.0 &   0.16 \\
\textbf{MNTDP-D* } &       0.64 &        0.64 &                  0.00 &               0.28 &             0.63 &             0.00 &  130.2 &   101.0 &   0.21 \\
\textbf{HAT*$\dagger$ } &       0.61 &        0.42 &                 -0.19 &               0.06 &             0.57 &            -0.01 &   26.6 &    54.0 &   0.12 \\
\textbf{HAT*$\dagger$ (Wide)} &       0.67 &        0.54 &                 -0.13 &               0.18 &             0.60 &             0.00 &  164.0 &   257.0 &   0.14 \\
\bottomrule
\end{tabular}

    \caption{\label{tab:res-remember-full} Results in the
      $\mathcal{T}(\mathcal{S}^{\mbox{\small{-}}})$ evaluation stream. In this stream, the last task
      is the same as the first with an order of magnitude less
      data. * correspond to models using an Alexnet backbone, $\dagger$ to models using stream-level cross-validation.}
\end{table}
\begin{figure}[h]	
	\centering
    \includegraphics[width=\textwidth]{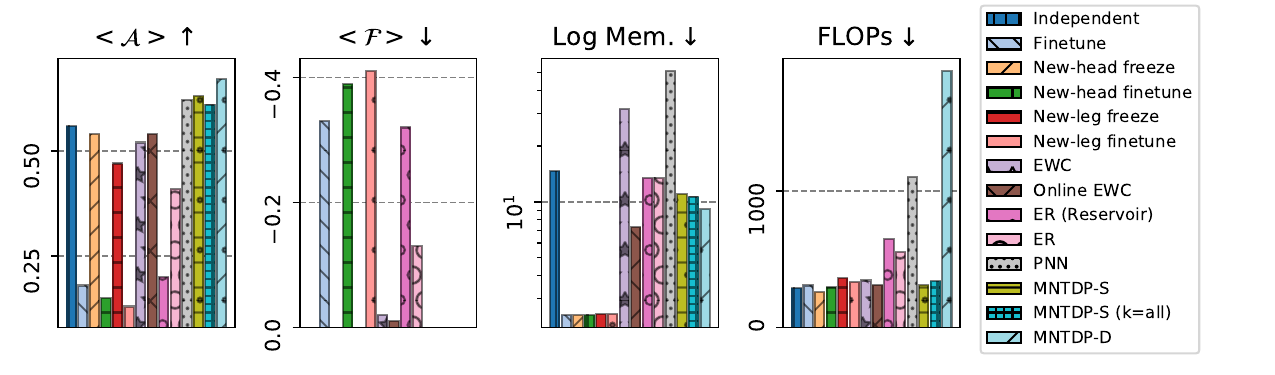}
	\caption{Comparison of all baselines on the $\mathcal{S}^{\mbox{\small{-}}}$
          stream. }\label{fig:fig:res-remember-barplot}
\end{figure}

\newpage
\subsection{Stream $\mathcal{S}^{\mbox{\small{+}}}$}

\begin{table}[h]
    \centering
    \begin{tabular}{lp{.8cm}p{.9cm}p{.8cm}p{.8cm}p{1cm}p{1cm}p{.8cm}p{.8cm}p{.8cm}}
\toprule
{} &  \mbox{Acc $T_1$} &  \mbox{Acc $T_1'$} &  $\Delta_{T_1, T_1'}$ &  $\mathcal{T}(\mathcal{S}^+)$ &  \mbox{$<\mathcal{A}>$} &  \mbox{$<\mathcal{F}>$} &   Mem. &   FLOPs &  LCA@5 \\
\textbf{Model             } &            &             &                       &                    &                  &                  &        &         &        \\
\midrule
\textbf{Independent       } &       0.37 &        0.71 &                  0.35 &               0.00 &             0.57 &             0.00 &   14.6 &   404.0 &   0.10 \\
\textbf{Finetune          } &       0.37 &        0.58 &                  0.21 &              -0.13 &             0.24 &            -0.23 &    2.4 &   262.0 &   0.12 \\
\textbf{New-head freeze   } &       0.37 &        0.43 &                  0.06 &              -0.28 &             0.41 &             0.00 &    2.5 &   361.0 &   0.17 \\
\textbf{New-head finetune } &       0.37 &        0.57 &                  0.20 &              -0.14 &             0.19 &            -0.36 &    2.5 &   327.0 &   0.11 \\
\textbf{New-leg freeze    } &       0.37 &        0.37 &                  0.01 &              -0.34 &             0.34 &             0.00 &    2.5 &   425.0 &   0.11 \\
\textbf{New-leg finetune  } &       0.37 &        0.56 &                  0.19 &              -0.15 &             0.17 &            -0.29 &    2.5 &   357.0 &   0.10 \\
\textbf{EWC  $\dagger$             } &       0.37 &        0.42 &                  0.05 &              -0.29 &             0.39 &            -0.02 &   31.5 &   337.0 &   0.11 \\
\textbf{Online EWC  $\dagger$      } &       0.37 &        0.40 &                  0.03 &              -0.31 &             0.37 &            -0.04 &    7.3 &   271.0 &   0.11 \\
\textbf{ER (Reservoir) $\dagger$   } &       0.37 &        0.56 &                  0.20 &              -0.15 &             0.20 &            -0.33 &   13.5 &   493.0 &   0.11 \\
\textbf{ER                } &       0.37 &        0.47 &                  0.10 &              -0.24 &             0.43 &            -0.07 &   13.5 &   512.0 &   0.12 \\
\textbf{PNN               } &       0.37 &        0.54 &                  0.17 &              -0.17 &             0.52 &             0.00 &   51.1 &  1757.0 &   0.12 \\
\textbf{MNTDP-S           } &       0.37 &        0.62 &                  0.25 &              -0.09 &             0.56 &             0.00 &   14.0 &   417.0 &   0.10 \\
\textbf{MNTDP-S (k=all)   } &       0.37 &        0.66 &                  0.29 &              -0.05 &             0.56 &             0.00 &   14.0 &   595.0 &   0.10 \\
\textbf{MNTDP-D           } &       0.37 &        0.72 &                  0.35 &               0.01 &             0.61 &             0.00 &   14.0 &  1659.0 &   0.11 \\
\textbf{MNTDP-D*} &       0.40 &        0.64 &                  0.24 &              -0.07 &             0.61 &             0.00 &  156.3 &   136.0 &   0.17 \\
\textbf{HAT*$\dagger$    } &       0.41 &        0.52 &                  0.12 &              -0.19 &             0.57 &             0.00 &   26.6 &    39.0 &   0.13 \\
\textbf{HAT*$\dagger$ (Wide)} &       0.41 &        0.61 &                  0.19 &              -0.10 &             0.59 &             0.00 &  164.0 &   324.0 &   0.14 \\
\bottomrule
\end{tabular}

    \caption{\label{tab:res-reset-full}Results in the $\mathcal{S}^{\mbox{\small{+}}}$ evaluation stream. In this stream, the 5th task is the same as the first with an order of magnitude more data. Tasks 2, 3, and 4 are distractors. * correspond to models using an Alexnet backbone, $\dagger$ to models using stream-level cross-validation.}
\end{table}

\begin{figure}[h]
	\centering
    \includegraphics[width=\textwidth]{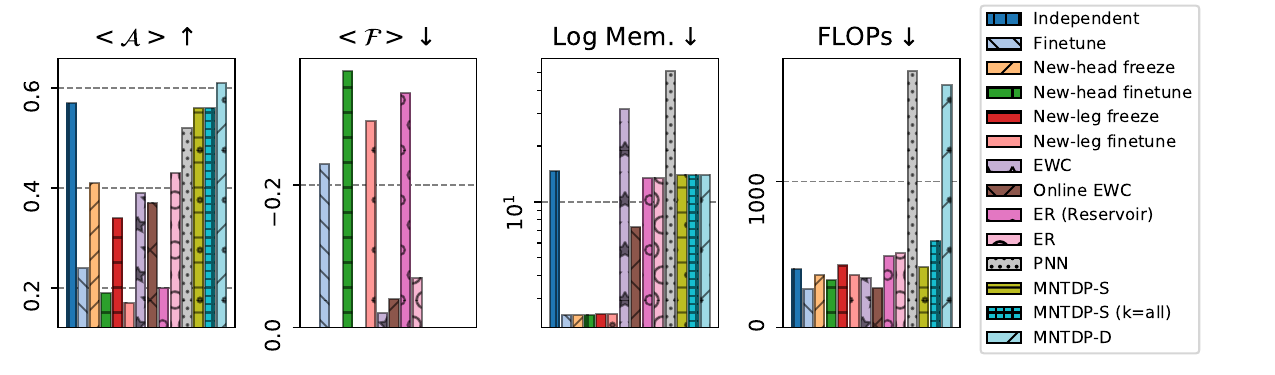}
	\caption{Comparison of all baselines on the $\mathcal{S}^{\mbox{\small{+}}}$ stream }\label{fig:res-reset-barplot}
\end{figure}

\newpage
\subsection{Stream $\mathcal{S}^{\mbox{\tiny{in}}}$}
\begin{table}[h]
    \centering
    \begin{tabular}{lp{.8cm}p{.9cm}p{.8cm}p{.8cm}p{1cm}p{1cm}p{.8cm}p{.8cm}p{.8cm}}
\toprule
{} &  \mbox{Acc $T_1$} &  \mbox{Acc $T_1'$} &  $\Delta_{T_1, T_1'}$ &  $\mathcal{T}(\mathcal{S}^{\mbox{in}})$ &  \mbox{$<\mathcal{A}>$} &  \mbox{$<\mathcal{F}>$} &   Mem. &   FLOPs &  LCA@5 \\
\textbf{Model             } &            &             &                       &                    &                  &                  &        &         &        \\
\midrule
\textbf{Independent       } &       0.98 &       0.60 &                 -0.39 &              0.00 &            0.57 &             0.00 &   14.6 &   265.0 &  0.10 \\
\textbf{Finetune          } &       0.98 &       0.57 &                 -0.41 &             -0.03 &            0.18 &            -0.31 &    2.4 &   244.0 &  0.11 \\
\textbf{New-head freeze   } &       0.98 &       0.39 &                 -0.59 &             -0.21 &            0.45 &             0.00 &    2.5 &   246.0 &  0.13 \\
\textbf{New-head finetune } &       0.98 &       0.62 &                 -0.36 &              0.02 &            0.19 &            -0.33 &    2.5 &   188.0 &  0.11 \\
\textbf{New-leg freeze    } &       0.98 &       0.95 &                 -0.04 &              0.35 &            0.48 &             0.00 &    2.5 &   282.0 &  0.10 \\
\textbf{New-leg finetune  } &       0.98 &       0.70 &                 -0.28 &              0.10 &            0.21 &            -0.34 &    2.5 &   238.0 &  0.10 \\
\textbf{EWC    $\dagger$           } &       0.98 &       0.87 &                 -0.12 &              0.27 &            0.43 &            -0.14 &   31.5 &   269.5 &  0.11 \\
\textbf{Online EWC $\dagger$       } &       0.98 &       0.87 &                 -0.12 &              0.27 &            0.43 &            -0.12 &    7.3 &   287.0 &  0.10 \\
\textbf{ER (Reservoir) $\dagger$   } &       0.98 &       0.60 &                 -0.38 &              0.00 &            0.30 &            -0.26 &   13.5 &   570.0 &  0.12 \\
\textbf{ER                } &       0.98 &       0.60 &                 -0.38 &              0.00 &            0.38 &            -0.17 &   13.5 &   604.0 &  0.12 \\
\textbf{PNN               } &       0.98 &       0.70 &                 -0.29 &              0.10 &            0.57 &             0.00 &   51.1 &   899.0 &  0.10 \\
\textbf{MNTDP-S           } &       0.98 &       0.64 &                 -0.34 &              0.04 &            0.57 &             0.00 &   12.2 &   333.0 &  0.10 \\
\textbf{MNTDP-S (k=all)   } &       0.98 &       0.68 &                 -0.30 &              0.08 &            0.57 &             0.00 &   13.4 &   327.0 &  0.10 \\
\textbf{MNTDP-D           } &       0.98 &       0.62 &                 -0.36 &              0.02 &            0.60 &             0.00 &   11.6 &  1225.0 &  0.11 \\
\textbf{MNTDP-D* } &       0.98 &       0.67 &                 -0.31 &              0.07 &            0.59 &             0.00 &  156.6 &   115.0 &  0.12 \\
\textbf{HAT*$\dagger$     } &       0.98 &       0.61 &                 -0.37 &              0.01 &            0.58 &            -0.01 &   26.6 &    41.0 &  0.12 \\
\textbf{HAT*$\dagger$ (Wide)} &       0.97 &       0.67 &                 -0.30 &              0.07 &            0.62 &             0.00 &  164.0 &   186.0 &  0.11 \\
\bottomrule
\end{tabular}
    \caption{\label{tab:res-leg-full}Results in the transfer evaluation stream with input perturbation. In this stream, the last task is the same as the first one with a modification applied to the input space and with an order of magnitude less data. Tasks 2, 3, 4 and 5 are distractors. * correspond to models using an Alexnet backbone, $\dagger$ to models using stream-level cross-validation.}
\end{table}
\begin{figure}[h]
	\centering
    \includegraphics[width=\textwidth]{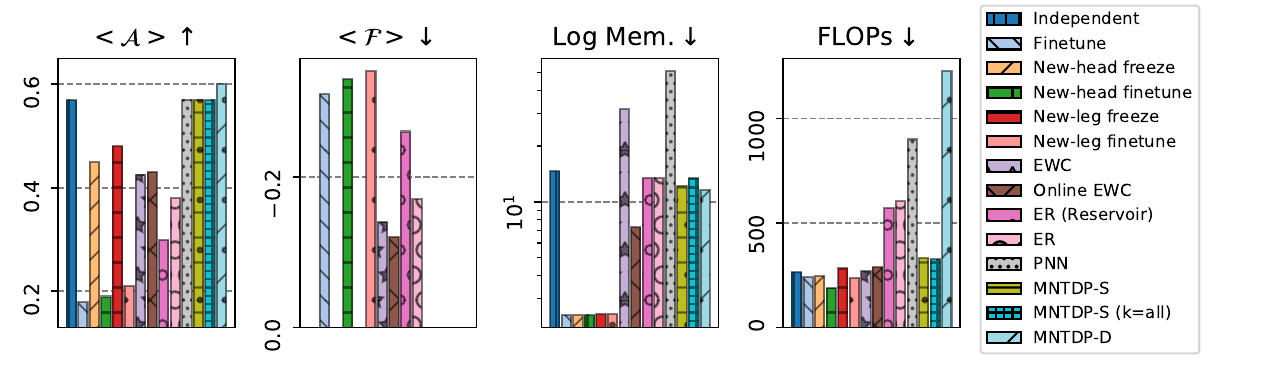}
	\caption{Comparison of all baselines on the $\mathcal{S}^{\mbox{\tiny{in}}}$ stream }\label{fig:res-leg-barplot}
\end{figure}

\newpage
\subsection{Stream $\mathcal{S}^{\mbox{\tiny{out}}}$}
\begin{table}[h]
    \centering
    \begin{tabular}{lp{.8cm}p{.9cm}p{.8cm}p{.8cm}p{1cm}p{1cm}p{.8cm}p{.8cm}p{.8cm}}
\toprule
{} &  \mbox{Acc $T_1$} &  \mbox{Acc $T_1'$} &  $\Delta_{T_1, T_1'}$ &  $\mathcal{T}(\mathcal{S}^{\mbox{out}})$ &  \mbox{$<\mathcal{A}>$} &  \mbox{$<\mathcal{F}>$} &   Mem. &   FLOPs &  LCA@5 \\
\textbf{Model             } &            &             &                       &                    &                  &                  &        &         &        \\
\midrule
\textbf{Independent       } &       0.70 &        0.37 &                 -0.32 &               0.00 &             0.61 &             0.00 &   14.6 &   349.0 &   0.10 \\
\textbf{Finetune          } &       0.70 &        0.36 &                 -0.33 &              -0.01 &             0.15 &            -0.37 &    2.4 &   331.0 &   0.11 \\
\textbf{New-head freeze   } &       0.70 &        0.70 &                  0.01 &               0.33 &             0.54 &             0.00 &    2.5 &   374.0 &   0.19 \\
\textbf{New-head finetune } &       0.70 &        0.31 &                 -0.39 &              -0.06 &             0.14 &            -0.41 &    2.5 &   379.0 &   0.10 \\
\textbf{New-leg freeze    } &       0.70 &        0.25 &                 -0.45 &              -0.12 &             0.40 &             0.00 &    2.5 &   369.0 &   0.10 \\
\textbf{New-leg finetune  } &       0.70 &        0.33 &                 -0.36 &              -0.04 &             0.14 &            -0.40 &    2.5 &   340.0 &   0.10 \\
\textbf{EWC   $\dagger$            } &       0.70 &        0.68 &                 -0.01 &               0.31 &             0.52 &            -0.03 &   31.5 &   387.0 &   0.11 \\
\textbf{Online EWC  $\dagger$      } &       0.70 &        0.66 &                 -0.04 &               0.29 &             0.51 &            -0.03 &    7.3 &   399.0 &   0.10 \\
\textbf{ER (Reservoir) $\dagger$   } &       0.70 &        0.39 &                 -0.31 &               0.02 &             0.22 &            -0.35 &   13.5 &   732.0 &   0.11 \\
\textbf{ER                } &       0.70 &        0.50 &                 -0.19 &               0.13 &             0.54 &            -0.07 &   13.5 &   758.0 &   0.11 \\
\textbf{PNN               } &       0.70 &        0.62 &                 -0.07 &               0.25 &             0.62 &             0.00 &   51.1 &  1799.0 &   0.10 \\
\textbf{MNTDP-S           } &       0.70 &        0.64 &                 -0.05 &               0.27 &             0.64 &             0.00 &   10.1 &   406.0 &   0.11 \\
\textbf{MNTDP-S (k=all)   } &       0.70 &        0.63 &                 -0.06 &               0.26 &             0.63 &             0.00 &   11.6 &   411.0 &   0.10 \\
\textbf{MNTDP-D           } &       0.70 &        0.70 &                  0.01 &               0.33 &             0.68 &             0.00 &   11.6 &  1299.0 &   0.15 \\
\textbf{MNTDP-D* } &       0.64 &        0.64 &                  0.00 &               0.27 &             0.65 &             0.00 &  130.2 &    99.0 &   0.22 \\
\textbf{HAT*$\dagger$     } &       0.62 &        0.44 &                 -0.18 &               0.07 &             0.60 &             0.00 &   26.6 &    42.0 &   0.13 \\
\textbf{HAT*$\dagger$ } &       0.68 &        0.51 &                 -0.17 &               0.14 &             0.64 &             0.00 &  164.0 &   293.0 &   0.12 \\
\bottomrule
\end{tabular}

    \caption{\label{tab:res-head-full}Results in the transfer evaluation stream with output perturbation. In this stream, the last task uses the same classes as the first task but in a different order and with an order of magnitude less data. Tasks 2, 3, 4 and 5 are distractors. * correspond to models using an Alexnet backbone, $\dagger$ to models using stream-level cross-validation.}
\end{table}
\begin{figure}[h]
	\centering
    \includegraphics[width=\textwidth]{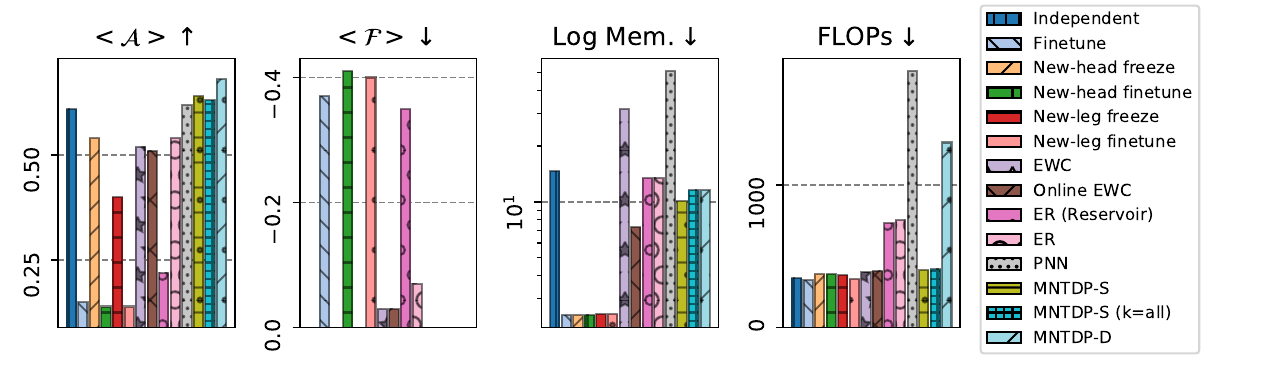}
	\caption{Comparison of all baselines on the $\mathcal{S}^{\mbox{\tiny{out}}}$ stream }\label{fig:res-hed-barplot}
\end{figure}

\newpage
\subsection{$\mathcal{S}^{\mbox{\tiny{pl}}}$}
\begin{table}[h]
    \centering
    \begin{tabular}{lp{1cm}p{1cm}p{1cm}p{1cm}p{1cm}p{1cm}p{1cm}}
\toprule
{} &  Acc $T_5$ &  $\Delta_{{T_5', T_5}}$ &  \mbox{$<\mathcal{A}>$} &  \mbox{$<\mathcal{F}>$} &   Mem. &   FLOPs &  LCA@5 \\

\textbf{Model             } &            &                         &                  &                  &        &         &        \\
\midrule
\textbf{Independent       } &       0.71 &                    0.00 &             0.59 &             0.00 &   12.2 &   232.0 &   0.10 \\
\textbf{Finetune          } &       0.57 &                   -0.14 &             0.21 &            -0.30 &    2.4 &   274.0 &   0.12 \\
\textbf{New-head freeze   } &       0.29 &                   -0.42 &             0.45 &             0.00 &    2.4 &   294.0 &   0.13 \\
\textbf{New-head finetune } &       0.56 &                   -0.15 &             0.20 &            -0.35 &    2.4 &   336.0 &   0.11 \\
\textbf{New-leg freeze    } &       0.27 &                   -0.44 &             0.37 &             0.00 &    2.5 &   390.0 &   0.11 \\
\textbf{New-leg finetune  } &       0.58 &                   -0.13 &             0.19 &            -0.34 &    2.5 &   375.0 &   0.11 \\
\textbf{EWC        $\dagger$       } &       0.28 &                   -0.43 &             0.27 &            -0.19 &   26.7 &   239.0 &   0.11 \\
\textbf{Online EWC   $\dagger$     } &       0.28 &                   -0.43 &             0.30 &            -0.17 &    7.3 &   282.0 &   0.12 \\
\textbf{ER (Reservoir)$\dagger$    } &       0.48 &                   -0.23 &             0.20 &            -0.27 &   11.7 &   383.0 &   0.10 \\
\textbf{ER                } &       0.51 &                   -0.20 &             0.45 &            -0.08 &   11.7 &   597.0 &   0.10 \\
\textbf{PNN               } &       0.56 &                   -0.15 &             0.54 &             0.00 &   36.5 &  1742.0 &   0.13 \\
\textbf{MNTDP-S           } &       0.58 &                   -0.13 &             0.55 &             0.00 &   11.0 &   351.0 &   0.10 \\
\textbf{MNTDP-S (k=all)   } &       0.60 &                   -0.11 &             0.56 &             0.00 &   11.0 &   340.0 &   0.11 \\
\textbf{MNTDP-D           } &       0.70 &                   -0.01 &             0.62 &             0.00 &   11.6 &  1503.0 &   0.10 \\
\textbf{MNTDP-D* } &       0.65 &                   -0.06 &             0.64 &             0.00 &  130.2 &   124.0 &   0.17 \\
\textbf{HAT*$\dagger$     } &       0.50 &                   -0.21 &             0.58 &             0.00 &   26.5 &    50.0 &   0.11 \\
\textbf{HAT*$\dagger$ (Wide)} &       0.61 &                   -0.10 &             0.61 &             0.00 &  163.7 &   312.0 &   0.12 \\
\bottomrule
\end{tabular}

    \caption{\label{tab:res-plasticity-full}Results in the plasticity evaluation stream. In this stream, we compare the performance on the probe task when it is the first problem encountered by the model and when it as already seen 4 distractor tasks.* correspond to models using an Alexnet backbone, $\dagger$ to models using stream-level cross-validation (see Section 5.2)}
\end{table} 
\begin{figure}[h]	
	\centering
    \includegraphics[width=\textwidth]{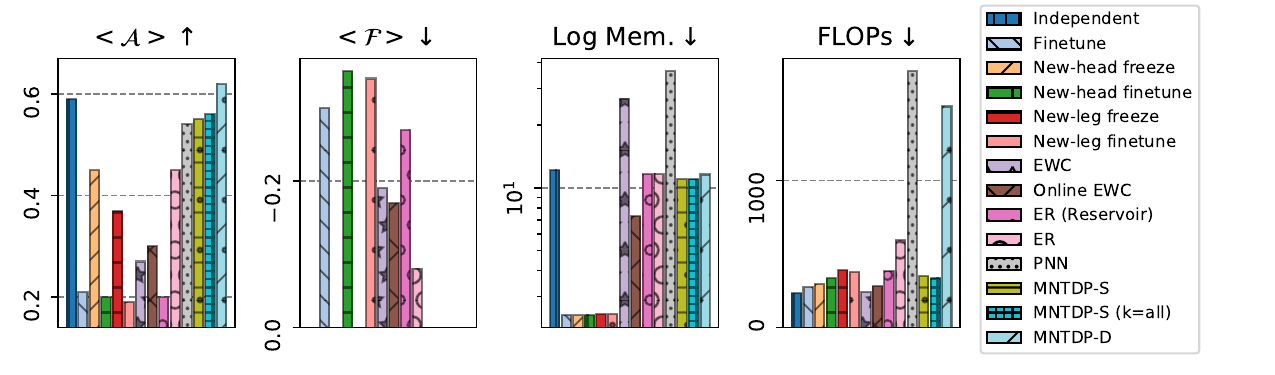}
	\caption{Comparison of all baselines on the $\mathcal{S}^{\mbox{\tiny{pl}}}$ stream }\label{fig:res-plasticity-barplot}
\end{figure}

\newpage
\subsection{$\mathcal{S}^{\mbox{\tiny{long}}}$}

\begin{table}[h]
    \centering



\begin{tabular}{lccccc}
\toprule
{} &   $<\mathcal{A}>$ &    $<\mathcal{F}>$ &                Mem. &                   FLOPs &            LCA@5 \\
Model              &                   &                    &                     &                         &                  \\
\midrule
\textbf{Independent}        &  0.57 $\pm$ 0.01  &     0.0 $\pm$ 0.0  &    243.7 $\pm$ 0.0  &   3542.33 $\pm$ 139.16  &   0.2 $\pm$ 0.0  \\
\textbf{Finetune }          &    0.2 $\pm$ 0.0  &   -0.35 $\pm$ 0.0  &      2.4 $\pm$ 0.0  &   4961.33 $\pm$ 112.16  &  0.23 $\pm$ 0.0  \\
\textbf{New-head freeze}    &  0.43 $\pm$ 0.01  &     0.0 $\pm$ 0.0  &      2.6 $\pm$ 0.0  &   5574.33 $\pm$ 249.65  &  0.27 $\pm$ 0.0  \\
\textbf{Online EWC}  $\dagger$       &  0.27 $\pm$ 0.01  &  -0.25 $\pm$ 0.01  &      7.4 $\pm$ 0.0  &   3882.67 $\pm$ 159.15  &  0.21 $\pm$ 0.0  \\
\textbf{MNTDP-S    }        &   0.68 $\pm$ 0.0  &     0.0 $\pm$ 0.0  &  158.63 $\pm$ 2.58  &   5437.67 $\pm$ 110.77  &  0.21 $\pm$ 0.0  \\
\textbf{MNTDP-D     }       &   0.75 $\pm$ 0.0  &     0.0 $\pm$ 0.0  &   102.03 $\pm$ 0.8  &  26066.67 $\pm$ 662.74  &  0.34 $\pm$ 0.0  \\
\textbf{MNTDP-D*}  &   0.75 $\pm$ 0.0  &     0.0 $\pm$ 0.0  &  1803.47 $\pm$ 16.45  &    2598.67 $\pm$ 70.48  &  0.46 $\pm$ 0.0  \\
\textbf{HAT*$\dagger$}     &  0.24 $\pm$ 0.01  &   -0.1 $\pm$ 0.03  &     31.9 $\pm$ 0.0  &      147.0 $\pm$ 27.39  &  0.21 $\pm$ 0.0  \\
\textbf{HAT (Wide) *$\dagger$} &   0.32 $\pm$ 0.0  &     0.0 $\pm$ 0.0  &    285.0 $\pm$ 0.0  &   1056.33 $\pm$ 137.29  &  0.21 $\pm$ 0.0  \\

\bottomrule
\end{tabular}

    \caption{\label{tab:res-long-stream-full}Results on the long evaluation stream. We report the mean and standard error using 3 different
    instances of the stream, all generated following the procedure described in \ref{app:datasets}.* correspond to models using an Alexnet backbone, $\dagger$ to models using stream-level cross-validation (see Section 5.2)}
\end{table} 

\begin{figure}[h]
	\centering
    \includegraphics[width=\textwidth]{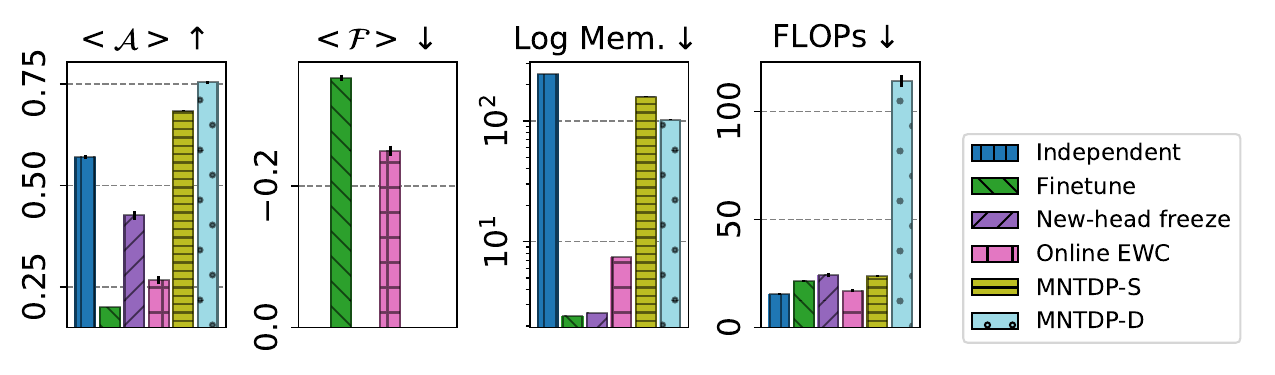}
	\caption{Comparison of all baselines on the $\mathcal{S}^{\mbox{\tiny{long}}}$ stream }\label{fig:bar_plot_long}
\end{figure}

\end{document}